\documentclass[11pt]{article}

\usepackage{NAACL2024}

\usepackage{times}
\usepackage{latexsym}

\usepackage[T1]{fontenc}

\usepackage[utf8]{inputenc}

\usepackage{microtype}

\usepackage{inconsolata}
\usepackage{graphicx}
\graphicspath{ {./diagrams/} }
\usepackage{array}
\usepackage{booktabs}
\usepackage{longtable}
\usepackage{multirow}
\usepackage[export]{adjustbox}
\usepackage{cellspace}
\usepackage{tabularray}
\usepackage[export]{adjustbox}
\usepackage{listings}
\usepackage{multicol}
\usepackage{pgfplots}
\usepgfplotslibrary{external}
\usepackage {algorithm}
\usepackage {algpseudocode}
\usepackage{amsmath}
\usepackage{courier}
\title{Leveraging Customer Feedback for Multi-modal Insight Extraction}

\author{Sandeep Sricharan Mukku\,, Abinesh Kanagarajan\,, Pushpendu Ghosh\,, Chetan Aggarwal \\ \\
        Amazon \\ 
        \\
\texttt{\{smukku, abinesk, gpushpen, caggar\}@amazon.com}}
\begin{document}
\maketitle
\begin{abstract}
Businesses can benefit from customer feedback in different modalities, such as text and images, to enhance their products and services. However, it is difficult to extract actionable and relevant pairs of text segments and images from customer feedback in a single pass. In this paper, we propose a novel multi-modal method that fuses image and text information in a latent space and decodes it to extract the relevant feedback segments using an image-text grounded text decoder. We also introduce a weakly-supervised data generation technique that produces training data for this task. We evaluate our model on unseen data and demonstrate that it can effectively mine actionable insights from multi-modal customer feedback, outperforming the existing baselines by $14$ points in F1 score. 

\end{abstract}
\section{Introduction}
\label{sec:introduction}
Customer feedback is essential for businesses to design and improve their products and services, according to customer expectations. \cite{info13070344} observe that the multi-modal feedback is growing rapidly. However, most existing solutions \cite{mukku2023insightnet, sircar2022distantly, liu2022leveraging} do not take into account the rich information that image and video feedback contain, which can enhance the actionability and improve the customer experience.  To address this challenge, we propose a novel multi-modal architecture that extracts pairs of text segments and corresponding images that are relevant and actionable for a given product from customer feedback. These pairs can help businesses to increase the actionability, improve the product catalogue quality, enhance the customer experience and thereby reduce returns and replacements.
\section{Related work}
\label{sec:related_work}
In recent years, multi-modal tasks, combining various data types such as images and text, have garnered significant interest in artificial intelligence and natural language processing~\cite{goyal2017making, zhou2020unified, lu2023multiscale}. Among these tasks, Visual Question Answering (VQA) stands out as a prominent domain~\cite{antol2015vqa}, wherein the aim is to generate textual answers to questions based on images. VQA has evolved significantly over the years, thanks to various advancing contributions. Early works on VQA~\cite{antol2015vqa} laid the groundwork, delineating the fundamental framework and challenges of the task. Subsequent research delved into innovative methodologies, such as leveraging transformer networks~\cite{zhou2020unified, lu2023multiscale}, and devising techniques for seamlessly integrating visual and textual information~\cite{li2019visualbert, lu2019vilbert, kim2021vilt}. The field progressed with more advanced models and datasets, such as VQAv2~\cite{goyal2017making}, which improved the robustness and performance benchmarking of VQA models. Moreover, the strides in pre-training on both vision and language have significantly influenced the landscape, as evidenced by groundbreaking models like OSCAR~\cite{li2020oscar}, BEiT-3~\cite{wang2022image}, and VLMo~\cite{bao2022beit, vimo2023vlmo}, which have achieved remarkable results. Additionally, techniques such as counterfactual data augmentation~\cite{chen2020counterfactual} have further enhanced the performance of VQA models.

Meanwhile, in the NLP domain, verbatim extraction and text summarization tasks have attracted a lot of attention. Models like BERT~\cite{devlin2018bert} and GPT-3~\cite{brown2020gpt3} have shown remarkable abilities in extracting and summarising textual feedback. Abstractive text summarization models, such as the T5~\cite{raffel2019t5}, have emerged as the state-of-the-art (SOTA) solutions for generating short and coherent summaries from longer texts.

Our work falls at the intersection of these two domains, where we propose a novel problem at the intersection of VQA and text summarization, where the input is a customer feedback text and corresponding images, and the objective is to extract the verbatim (exact segments of the textual feedback) that best highlights or talks about the feedback image. We design a multi-modal approach that builds on VQA models and extends them with text generation capabilities, introducing a new technique for multi-modal insight extraction. Our work bridges the gap between multi-modal understanding and verbatim extraction, and enhances the interpretability of images within textual context, especially for customer feedback and images.

\section{Problem Statement}
\label{sec:problem_statement}
Given a customer feedback consisting of a set of images $I = \{I_1, I_2, ... I_n\}$ and a text $T$, we extract $m$ verbatims from $T$, denoted by $V = \{v_i\ |\ 1\leq~i~\leq~m, v_i \in T\}$. The aim is to extract all the relevant and actionable verbatims $\{v_k \in V\}$ that corresponds to the given image $I_k$. 
\section{Proposed Approach}
\label{sec:methedology}
In this section, we first present our data generation method that creates training data from raw feedback. Then, we introduce our model architecture that extracts pairs of verbatims and images that are relevant and actionable for a given product.

\begin{figure*}
\centering 
\includegraphics[width=\textwidth]{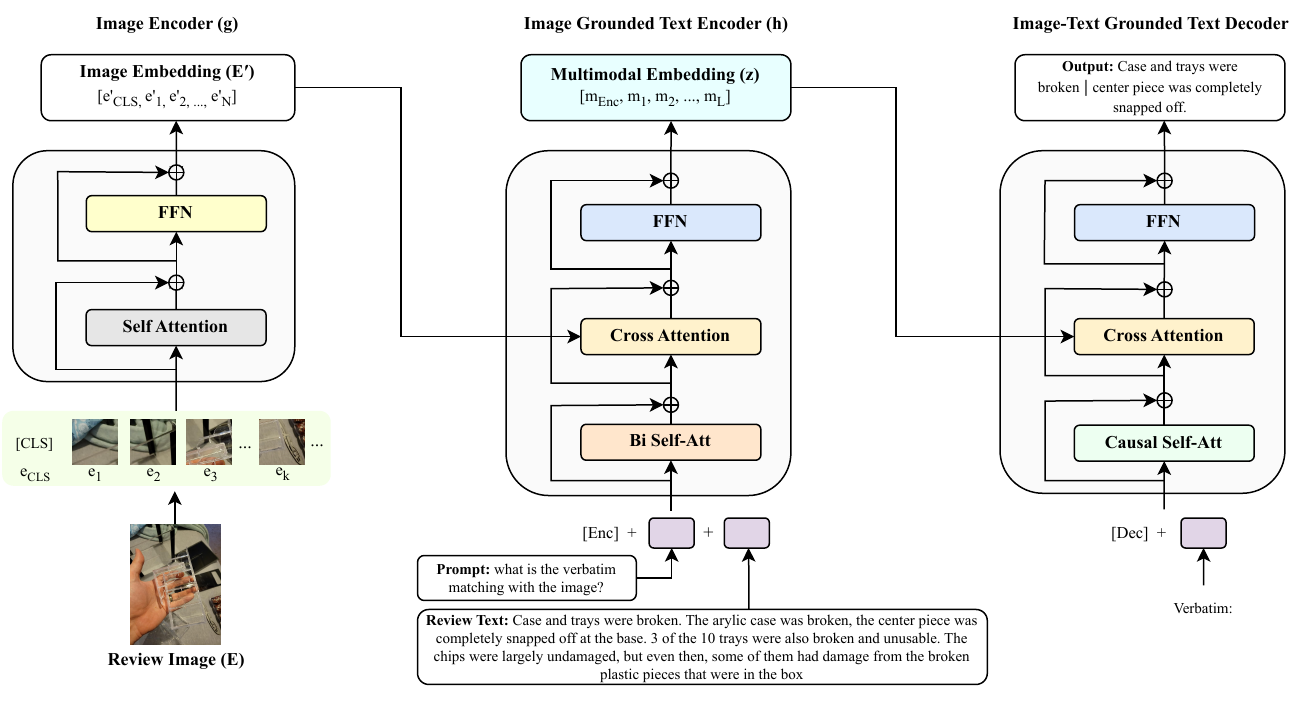} 
\caption{\underline{M}ulti-modal \underline{IN}sight \underline{E}xtraction (MINE) Architecture} 
\label{fig:model_architecture} 
\end{figure*}

\subsection{Weakly supervised Training Data Generation}
\label{sec:data_generation_mechanism}
We segment raw feedback text to extract actionable verbatim\footnote{key text-phrases extracted from customer feedback, that can be utilised to take actions} (refer appendix section~\ref{sec:verbatim_extraction} for exact process). We then generate training data by obtaining $m \times n$ possible verbatim-image pairs for each feedback text, where $m$ is the number of verbatim and $n$ is the number of images. Next, we compute cosine similarity scores for every verbatim-image pair using pre-trained CLIP~\cite{radford2021learning}. To evaluate the relevance of the feedback image-verbatim pairs predicted by base CLIP\footnote{We used the \href{https://huggingface.co/docs/transformers/model_doc/clip}{ViT-B/32} version of CLIP}, we had them manually annotated as positive or negative pairs with clearly defined annotation guidelines (refer section~\ref{sec:annotation_guidelines_insight_relevancy}). We established a validation set, denoted as \textbf{StratSet-1k}, comprising $1,000$ image-verbatim pairs, carefully stratified (Algorithm~\ref{alg:stratified_sample}) across $27$ distinct product categories sourced from the raw dataset by clustering\footnote{We used \href{https://github.com/UKPLab/sentence-transformers/blob/master/examples/applications/clustering/fast_clustering.py}{Fast-clustering} to cluster the verbatims}\textsuperscript{,}\footnote{Clustering verbatims helps group the relevant intents and thereby diversify the sample set} the actionable verbatims. 

\begin{algorithm}[H]
\caption{Stratified Sampling(K)}
\label{alg:stratified_sample}
\small
\begin{algorithmic}[1]
\State Let $C$ be the number of clusters of verbatims spread across $L$ product categories.
\State Let $K$ be the no. of verbatims to be sampled
\State $S \gets []$
\For{$c \in C$}
    \State $V_c \gets \text{verbatims in cluster } c$
    \State $D_c \gets \{\text{product categories in } V_c\}$
    \For{$d \in D_c$}
        \State $V_{c,d} \gets \text{verbatims in category } d \text{ of cluster } c$
        \State $k_d \gets \lfloor K \cdot \frac{1}{|L|} \rfloor$
        \State $S \gets S \cup \text{RandomSample}(V_{c,d}, k_d)$
    \EndFor
\EndFor
\State \textbf{return} $S$

\end{algorithmic}
\end{algorithm}

To determine the optimal threshold, we systematically adjusted it within the range of $0.19$ to $0.31$ and recorded the resulting precision and product category coverage. Our analysis (refer Figure~\ref{fig:training_data_to_train_data_creation_model}) revealed that a threshold of $0.27$ strikes a favorable balance between product category coverage and precision, effectively reducing noise in the training data. We prioritized precision, aiming for a minimum of $90\%$, which sufficiently covers the majority of product categories while accepting a lower level of recall, as these matched pairs are primarily used for fine-tuning the CLIP. 

\begin{figure}[H]
\centering
\begin{tikzpicture}
\begin{axis}[
    width=8cm,
    height=8cm,
    legend style={font=\small},
    xlabel={Image-verbatim similarity score on \textbf{StratSet-1k}},
    ylabel={},
    xmin=0.18, xmax=0.33,
    ymin=0.45, ymax=1.05,
    xtick={0.19, 0.21, 0.23, 0.25, 0.27, 0.29, 0.31},
    ytick={0.5, 0.6, 0.7, 0.8, 0.9, 1},
    legend pos=south west,
    grid style=dashed,
]
\addplot[
    color=blue,
    mark=square,
    thick
    ]
    coordinates {
    {(0.19,0.74)(0.2,0.74) (0.21,0.76) (0.22,0.79) (0.23,0.83) (0.24,0.87) (0.25,0.89) (0.26,0.9)  (0.27,0.93) (0.28,0.93) (0.29,0.93) (0.3,0.95) (0.31,1)}
    };
    \addlegendentry{Precision};

\addplot[
    color=red,
    mark=o,
    thick
    ]
    coordinates {
    {(0.19,0.99)(0.2,0.98) (0.21,0.97) (0.22,0.96) (0.23,0.95) (0.24,0.92) (0.25,0.89) (0.26,0.84)  (0.27,0.81) (0.28,0.74) (0.29,0.68) (0.3,0.61) (0.31,0.5)}
    };
    \addlegendentry{Category Coverage};
\addplot [black, dashed, thick] coordinates {(0.27,0.4) (0.27,1)};
    \addlegendentry{Threshold};
    
\end{axis}
\end{tikzpicture}
\caption{Product category coverage and precision as a function of raw data threshold for fine-tuning CLIP. The graph shows the trade-off between coverage and precision for different values of the threshold}
\label{fig:training_data_to_train_data_creation_model}
\end{figure}

After that, we fine-tune the CLIP using $210K$ positive pairs with a symmetric cross-entropy loss (refer Eqn.~\ref{eq:clip_symmetric_loss}) that aligns image and text embeddings. The scaled pairwise cosine similarities between image and text embeddings are computed as follows:



\begin{multline}
logits_{I,T} = \frac{e_I^\text{T} e_T}{|e_I| |e_T|} / \tau
= \cos(\theta_{I,T}) / \tau
\end{multline}
where $e_I$ and $e_T$ are the image and text embeddings, and $\theta_{I,T}$ is the angle between them. We compute the loss over both images and verbatim (text) by averaging the losses for each modality:

\begin{multline}
\label{eq:clip_symmetric_loss}
loss = (loss_i + loss_t) / 2 \\
= -\frac{1}{2} \sum_{i=1}^n \left[ \log \frac{\exp(\cos(\theta_{i,labels_i}) / \tau)}{\sum_{t=1}^m \exp(\cos(\theta_{i,t}) / \tau)} \right. \\
+ \left. \log \frac{\exp(\cos(\theta_{labels_i,i}) / \tau)}{\sum_{j=1}^n \exp(\cos(\theta_{j,i}) / \tau)} \right]
\end{multline} 
where $labels_i \in V$ is the most relevant text of the $i$-th image, $j$ iterates over the image classes, and $\tau$ is a temperature parameter that controls the smoothness of softmax distribution. This loss function encourages the model to learn embeddings that are close to each other for positive pairs and far apart for negative pairs. The loss is symmetric as it is computed over both modalities. Finally, we use this trained model to obtain the matched (positive) and mismatched (negative) pairs from segments and images along with their similarity scores (sample training example shown in appendix section~\ref{sec:training_data}).

\begin{figure}[H]
\centering
\begin{tikzpicture}
\begin{axis}[
    width=8cm,
    height=8cm,
    legend style={font=\small},
    xlabel={Image-verbatim similarity score on \textbf{StratSet-1k}},
    ylabel={},
    xmin=0.18, xmax=0.3,
    ymin=0.05, ymax=1.05,
    xtick={0.19, 0.21, 0.23, 0.25, 0.27, 0.29},
    ytick={0, 0.2, 0.4,  0.6,  0.8,  1},
    legend pos=south west,
    grid style=dashed,
    title={},
]
\addplot[
    color=blue,
    mark=square,
    thick
    ]
    coordinates {
    {(0.19, 0.74)
(0.2, 0.74)
(0.21, 0.76)
(0.22, 0.79)
(0.23, 0.83)
(0.24, 0.87)
(0.25, 0.89)
(0.26, 0.92)
(0.27, 0.91)
(0.28, 0.94)
(0.29, 0.94)
}
    };
    \addlegendentry{Precision};

\addplot[
    color=red,
    mark=o,
    thick
    ]
    coordinates {
    {(0.19, 0.99)
(0.2, 0.98)
(0.21, 0.95)
(0.22, 0.89)
(0.23, 0.79)
(0.24, 0.65)
(0.25, 0.53)
(0.26, 0.41)
(0.27, 0.29)
(0.28, 0.2)
(0.29, 0.11)
}
    };
    \addlegendentry{Recall};
\addplot[
    color=green,
    mark=diamond,
    thick
    ]
    coordinates {
    {(0.19, 0.85)
(0.2, 0.84)
(0.21, 0.85)
(0.22, 0.84)
(0.23, 0.81)
(0.24, 0.74)
(0.25, 0.66)
(0.26, 0.56)
(0.27, 0.44)
(0.28, 0.32)
(0.29, 0.2)
}
    };
     \addlegendentry{F1 Score};
\addplot [black, dashed, thick] coordinates {(0.225,0.4) (0.225,1)};
    \addlegendentry{Threshold};
\end{axis}
\end{tikzpicture}
\caption{Training data threshold selection}
\label{fig:training_data_threshold_selection_actual_model}
\end{figure}

We employed the fine-tuned CLIP model to make inferences on the StratSet-1k dataset. Subsequently, we manually annotated the inferred pairs as either positive or negative, adhering to  Inter-Annotator Agreement (IAA) process~\cite{artstein2017inter}. We then computed Precision, Recall, and F1-score across a range of threshold values, as illustrated in Figure~\ref{fig:training_data_to_train_data_creation_model}. Our analysis revealed that a threshold of $0.225$ offers an optimal balance between precision and recall, thus serving as the foundation for creating training data for our problem. This refinement resulted in a notable enhancement of the fine-tuned CLIP model, achieving $75\%$ F1-score, representing a $3\%$ improvement compared to the off-the-shelf CLIP model.

\subsubsection{Annotation Guidelines: Verbatim-Image relevancy}
\label{sec:annotation_guidelines_insight_relevancy}
Following are the annotations guidelines given to annotators that define the relevancy of Verbatim-Image pair:

\begin{enumerate}
    \item \textbf{Object relevance}: When certain object discussed in the text is found in the image in any forms which is may not be explicitly mentioned, it will be considered as relevant pair.
    \item \textbf{OCR relevance}: When certain entity discussed in the text is represented in the image by some form of text, the pair is considered as relevant. 
    \item \textbf{Semantic relevance}: When the information discussed in the text is contextually represented in the image, the pair to be marked relevant. 
\end{enumerate}

Each pair is annotated (\textit{relevant} / \textit{not relevant}) by two experts separately and resolved by third in-case of conflicting annotation. We used IAA process and achieved Cohen's kappa~\cite{cohen1960coefficient} of $0.89$. 

\subsection{Model Components}
\label{sec:model_components}
In this section, we describe the components of our proposed architecture for multi-modal insight extraction, which we refer to as MINE. Motivated by BLIP~\cite{li2022blip}, MINE consists of three main modules: an image encoder, an image-grounded text encoder, and an image-text grounded text decoder. The image encoder uses a visual transformer to extract visual features from the input image. The image-grounded text encoder fuses the input text with the visual features using cross-attention layers. The image-text grounded text decoder generates the output text using causal self-attention and cross-attention layers conditioned on the multi-modal representation from the encoder and the previous tokens.

\subsubsection{Image Encoder}
\label{sec:image_encoder}
Image encoder is built using the visual transformer ViT-B/16 \cite{dosovitskiy2020vit}, which consists of an image encoder and a transformer encoder. The image encoder splits the input image into patches of size $16 \times 16$ pixels and converts each patch into a vector of $768$ dimension embedding. The transformer encoder (which has $12$ layers and each layer performs different operations on the embeddings, such as attention, normalization, and feed-forward networks) takes the sequence of embeddings and outputs a new sequence of embeddings that contains more information about the image content and context. We formulate the transformer encoder as a function $g$ that maps a sequence of embeddings $E$ to another sequence of embeddings $E'$:

\begin{equation}
E' = g(E) = [e'_{\text{CLS}}, e'_1, e'_2, ..., e'_N]
\end{equation}
where $e'_{\text{CLS}}$ is the updated [CLS] token (added to represent the embeddings as image features), $e'_k$ is the updated embedding for the $k$-th patch ($e_k$), and $N$ is the number of patches in the image.

\subsubsection{Image Grounded Text Encoder}
\label{sec:image_grounded_text_encoder}
The text encoder is based on BERT base \cite{devlin2018bert}, which has $12$ transformer blocks with self-attention and feed-forward network (FFN) layers. To fuse the visual embedding from the image encoder, an extra cross-attention layer is inserted between the self-attention and FFN layers in each transformer block. This layer updates the text embeddings by attending to the image embeddings. The attention mechanism in the cross-attention layer computes attention scores for each token in the text sequence with respect to the image embeddings and determine how much importance each token in the text should place on the information in the image embeddings. A special token [ENC] is appended to the start of the input text (refer Section \ref{sec:prompt_engineering} for exact prompts used) to provide an identity for the encoder input. The text encoder outputs a multi-modal embedding of the image-text pair as follows: 

\begin{equation}z = h(T, E') = [m_{\text{ENC}}, m_1, m_2, ..., m_L] \end{equation}
where $h$ is the image-grounded text encoder function, $T$ is the text input, $E'$ is the embedding from image encoder ($g$), $m_{\text{ENC}}$ is the multi-modal embedding for the [ENC] token, $m_i$ is the multi-modal embedding for the $i$-th text token, and $L$ is the length of the text input.

\begin{table*}
\centering
\begin{tabular}{cccccc}
\hline
\textbf{Model} &
  \textbf{Prompting Approach} &
  \multicolumn{1}{l}{\textbf{Precision (Correctness)}} &
  \multicolumn{1}{l}{\textbf{Recall}} &
  \multicolumn{1}{l}{\textbf{F1-score}} &
  \multicolumn{1}{l}{\textbf{Completeness}} \\ \hline
\multirow{4}{*}{\textbf{ALBEF}} & CSECS  & 0.61	& 0.52	& 0.56	& 0.71     \\
                                & MSECS  & 0.63	& 0.59	& 0.61	& 0.76     \\
                                & MSE  & 0.65	& 0.60	& 0.62	& 0.80     \\
\hline
\multirow{3}{*}{\textbf{VL-T5}} & CSECS  & 0.65	& 0.53	& 0.58	& 0.76     \\
                                & MSECS  & 0.65	& 0.57	& 0.61	& 0.82     \\
                                & MSE  & 0.66	& 0.61	& 0.63	& 0.89     \\
                                \hline
\multirow{3}{*}{\textbf{MINE}} & CSECS  & 0.69	& 0.70	& 0.69	& 0.73     \\
                                & MSECS  & 0.71	& 0.76	& 0.73	& 0.91     \\
                                & MSE   & \textbf{0.76}	& \textbf{0.77}	& \textbf{0.77}	& \textbf{0.93}   \\
                                \hline

\end{tabular}
\caption{Experimental results}
\label{tab:experimental_results}
\end{table*}

\subsubsection{Image-Text Grounded Text Decoder}
\label{sec:image_text_grounded_text_decoder}
The decoder follows the same architecture as the encoder, except that it uses causal self-attention instead of bidirectional self-attention. The multi-modal embedding is also fused as a cross-attention layer between FFN and attention layer. The decoder shares the parameters of FFN and cross-attention layers with the encoder, which improves training efficiency and enables multitask learning. The causal self-attention layer allows the decoder to generate text tokens ($y_t$), conditioned on the previous tokens ($y_{<t}$) and the multi-modal representation ($z$). For a given input token ($x_t$), the output of the decoder at each time step $t$ is:

\begin{equation}
    y_t = f(x_t, y_{<t}, z)
\end{equation}

\subsection{Overall Architecture}
\label{sec:overall_architecture}
We encode the review text and image pair into a multi-modal representation and decode it into a sequence of verbatims (feedback segments) that are relevant to the image. Figure~\ref{fig:model_architecture} illustrates the overall architecture of MINE. The encoding process consists of two steps: we apply the image encoder to the review image to obtain an image embedding; then, we pass the review text and the image embedding to the image-grounded text encoder to produce a multi-modal embedding that fuses both modalities. The decoding process uses the image-text grounded text decoder, which takes the multi-modal embedding as input and generates verbatim tokens conditioned on it. During training, we provide the ground truth verbatim as input to the decoder and optimize it to predict the next token. We fine-tune our model, with an objective of extracting the insightful segment by optimizing cross entropy loss:
\begin{multline*}
L(\theta) = -\frac{1}{N}\sum_{n=1}^{N}\sum_{t=1}^{T_n}\log p(y_t^n|y_{<t}^n,x^n,z^n;\theta) \, (6)
\end{multline*}
where $N$ is the number of review-image pairs in the dataset, $T_n$ is the length of the verbatim sequence for the $n$-th pair, $y_t^n$ is the verbatim token at time step $t$ for the $n$-th pair, $y_{<t}^n$ is the sequence of previous tokens, $x^n$ is the review verbatim, $z^n$ is the multi-modal embeddings from the image grounded text encoder, and $\theta$ are the model parameters. During inference, we only give the [DEC] token as input to the decoder and let it generate verbatims.

\section{Experimental settings}
\label{sec:experiments}
\subsection{Dataset \& Analysis}
\label{sec:dataset-and-analysis}
We obtain raw reviews dataset\footnote{dataset can be found \href{https://nijianmo.github.io/amazon/index.html}{here} \label{amazon_us_reviews}} provided by~\cite{ni-etal-2019-justifying}, which contains over $233$ million reviews from $29$ unique product categories. We used the sample subset (K-cores subset, released as part of the original dataset) which have $973$k reviews with images of $27$ product categories for our analysis.

\subsection{Prompt Engineering}
\label{sec:prompt_engineering}
We use the similarity scores that are obtained from the training data generation, as the confidence scores for each verbatim (see appendix section~\ref{sec:training_data} for an example). We compare three different prompting approaches for extracting actionable verbatim from the review text and image pair, and are as follows:

\vspace{-0.2cm}

\begin{enumerate}
    \item \textbf{Comprehensive Segment Extraction with Confidence Scores (CSECS)}: We generate all possible verbatim from the text and their scores based on how well they match the image. The ground truth includes all verbatim and their scores for each pair.
    
    \vspace{-0.2cm}
    
    \item \textbf{Matching Segment Extraction with Confidence Scores (MSECS)}: We generate only the verbatim that match image and their scores. The ground truth includes only matching verbatim and their scores for each pair.
    
    \vspace{-0.2cm}
    
    \item \textbf{Matching Segment Extraction (MSE)}: We generate only the verbatim that match the image without any scores. The ground truth includes only the matching verbatim predicted by the fine-tuned CLIP model for each pair.
\end{enumerate}

\vspace{-0.3cm}
Refer appendix section~\ref{sec:experimented_prompts} for exact prompts and targets used during training.

\subsection{Approaches}
\label{sec:approaches}
As an additional multi-modal approaches, we tried ALBEF~\cite{li2021align} and VL-T5~\cite{cho2021unifying} along with MINE. We use $80K$ verbatim-image pairs sampled from all product categories to train both approaches.

\textbf{ALBEF}: Given that the training data was generated using a weakly supervised methodology and may contain noise, we employ ALBEF, a state-of-the-art robust multi-modal vision and language representation learning model. ALBEF leverages momentum distillation, a self-training technique, to glean knowledge from pseudo-targets, enhancing its resilience to noise within the training data. We considered the VQA setting\footnote{\url{https://github.com/salesforce/ALBEF/blob/main/VQA.py}} for our baselines. ALBEF consists of an image encoder and a text encoder, followed by a 6-layer auto-regressive answer decoder, where we used the pretrained encoder weights and only finetuned the decoder for all the three prompting approaches, as mentioned in section~\ref{sec:prompt_engineering}.  

\textbf{VL-T5}: We fine-tune VL-T5, on our task of generating actionable verbatim from review text and image. We use Faster R-CNN~\cite{ren2016faster} to obtain $36$ object features from the review image, which is concatenated with the tokenized input text consisting of the prompt question and the review text. The resulting sequence is fed into the bidirectional multi-modal encoder of VL-T5, which learns to encode both textual and visual information. The decoder of VL-T5 then generates the actionable verbatim as the output.


\subsection{Training Details}
\label{sec:hyperparameters}
MINE was initialized using the pre-trained weights of BLIP base model and fine-tuned using AdamW optimizer~\cite{loshchilov2019decoupled} for $10$ epochs with a cosine learning rate scheduler~\cite{loshchilov2017sgdr} and a minimum and initial learning rate of $1e^{-6}$ and $5e^{-5}$ respectively. In comparison, ALBEF was fine-tuned using $128$ as batch size for $20$ epochs with the following hyperparameters: AdamW optimizer, a learning rate of $1e^{-5}$, $\beta_1 = 0.9$, $\beta_2 = 0.999$, $\epsilon = 10^{-8}$ and a weight decay of $0.05$. Similarly, we fine-tuned VL-T5 for $5$ epochs using AdamW optimiser with a learning rate of $5e^{-5}$ and linear warmup scheduler of $5\%$. At inference, we adjusted the max length to $512$ to enable the model to generate longer and more relevant verbatim. Additionally, we employed an image resolution of $225$px, consistent with the BLIP base model. We also experimented with different decoding methods for MINE, such as beam search~\cite{hu-etal-2015-improved} ($beam\_size=10$), consistent top-k sampling~\cite{welleck2020consistency} ($top\_k=50$) and nucleus sampling~\cite{holtzman2019curious} ($top\_k=50$ and $top\_p=0.95$).
\subsection{Results}
We found that MSE approach with beam search decoding produced the most complete and relevant verbatim. We compared different prompt settings and approaches, and measured the quality of the generated verbatim using precision (correctness), recall and completeness metrics. We gave the exact definitions of these metrics in appendix section~\ref{sec:verbatim_evaluation_metrics}. Table~\ref{tab:experimental_results} summarized the results of our experiments. We see that our approach is effective than existing baselines like ALBEF and VL-T5 in both identifying the exact intent and extracting the complete verbatim, present in the image. In addition to beam search, we experimented with different decoding methods for MINE and reported the results in Table~\ref{tab:additional_experimental_results}.

\begin{table*}
\centering
\begin{tabular}{cccccc}
\hline
\textbf{Prompt Type} &
  \textbf{Decode method} &
  \multicolumn{1}{l}{\textbf{Precsion (Correctness)}} &
  \multicolumn{1}{l}{\textbf{Recall}} &
  \multicolumn{1}{l}{\textbf{F1-score}} &
  \multicolumn{1}{l}{\textbf{Completeness}} \\ \hline
\multirow{2}{*}{\textbf{CSECS}} & topK        & 0.57  &	0.48  &	 0.52  &	0.66 \\
                                & nucleus     & 0.62 & 	0.6	& 0.61 &	0.71  \\ 
                                \hline
\multirow{2}{*}{\textbf{MSECS}} & topK        & 0.6	& 0.59	& 0.59	& 0.69  \\
                                & nucleus     & 0.65 &	0.69 &	0.67 &	0.83  \\ 
                                \hline
\multirow{2}{*}{\textbf{MSE}}   & topK        & 0.57 &	0.7	& 0.63 & 0.72  \\
                                & nucleus     & 0.68 &	0.72 &	0.7	& 0.86 \\ 
                                \hline
\end{tabular}
\caption{Additional Experimental results}
\label{tab:additional_experimental_results}

\end{table*}

\section{Conclusion}
\label{sec:conclusion}
In this paper, we introduced a novel architecture MINE, for extracting insights from multi-modal customer feedback encompassing both text and images. We proposed a weakly-supervised data generation approach that leveraged raw feedback text and images to create relevant verbatim and image pairs. We also proposed an unsupervised insight extraction approach that used an image-grounded text encoder to learn latent representations of the feedback text and corresponding images, and an image-text grounded text decoder to extract actionable verbatim relevant to image. We evaluated our approach on real-world dataset and demonstrated its effectiveness in extracting actionable insights from multi-modal feedback with minimal supervision, achieving a 14-point improvement in F1-score over the existing baselines and uni-modal insight extraction.

\textbf{Future work:} Our research opens up several avenues for further exploration. As a future direction, we would like to extend our method to video feedback by incorporating the temporal dimension and extracting relevant video snippets, as mention in the feedback text. We also aim to structure the extracted insights in hierarchical way that facilitates decision making for users. Moreover, we plan to localise the regions of interest in the feedback images that correspond to the verbatim, using bounding boxes, to help identify the specific features or issues that customers refer to in their feedback. Finally, we intend to generate suggestions for product enhancements or fixes based on the extracted insights and the customer sentiment.

\section*{Limitations}
\label{sec:Limitations}
Like any other ML architecture, even MINE architecture has some limitations. One of the limitations is that the MINE architecture is not pre-trained with language modelling as an objective (such as Prefix language modelling-style / Deshuffling-style / i.i.d noise, replace spans-style~\cite{raffel2020exploring}, or BERT-style~\cite{devlin2019bert}, or MASS-style~\cite{song2019mass}). This means that the model may not always generate responses that are identical to the feedback texts in terms of wording, even though they generate text the same intent. We found that around $19$\% of the cases had different wording but with same intent. Another limitation is that the extraction task can be computationally demanding with a higher number of beams, which may affect the performance and scalability of the approach using low compute resources.
\section*{Ethics Statement}
\label{sec:ethics_statement}
In this paper, we used a publicly available dataset in an almost unsupervised fashion, with minimal reliance on human annotations for labeling. Annotators are subject matter experts who specialize in annotating image-text samples, whether they are relevant or not. They are compensated accordingly, following industry standards set by the organisation. We carefully considered all aspects related to annotator agreement and fairness. However, the dataset is anonymized and does not contain any identifiable or traceable information. Thus, we respect the privacy and confidentiality of the consumers of the product and do not expose them to any potential harm or misuse. The dataset is widely used and well-known in the natural language processing community. It has been previously analyzed and evaluated by several researchers and practitioners. Our work does not introduce any bias or prejudice either, as we do not make any assumptions or judgments based on the feedback text or images. Our work maintains a purely empirical and objective approach, without favoring or disadvantaging any individual or product. Throughout our research process and reporting, we adhered to the ACL code of ethics and professional conduct.
\bibliography{anthology,custom}
\bibliographystyle{acl_natbib}
\appendix
\twocolumn 
\section{Appendix}
\label{sec:appendix}

\subsection{Verbatim Extraction}
\label{sec:verbatim_extraction}
We use the following steps to extract verbatims (actionable feedback text segments) from raw customer feedback source:

\begin{enumerate} 

\item \textbf{Text Pre-processing:} We apply text pre-processing techniques to remove html tags, urls, accented characters, extra spaces and other noise from the feedback text. 

\item \textbf{Segmentation:} We use Trankit~\cite{nguyen2021trankit}, a fast and lightweight transformer-based toolkit for natural language processing, to segment each feedback into meaningful units. We also implement some heuristics to avoid having segments with only single word. 

\item \textbf{Removing non-actionable (neutral) segments:}We use a RoBERTa~\cite{liu2019roberta} based 3-class sentiment classifier\footnote{\url{https://huggingface.co/siebert/sentiment-roberta-large-english}} to evaluate the sentiment of each feedback segment. It has been rigorously evaluated and demonstrates an impressive 92\% F1-score on our specific sentiment classification task. We make the decision to exclude segments classified as neutral, as these segments may not provide actionable feedback for brands and selling partners who seek feedback to improve their products and services.
\end{enumerate}

\subsection{Verbatim evaluation metrics}
\label{sec:verbatim_evaluation_metrics}
\begin{enumerate}
    \item \textbf{Completeness}: This metric measures the proportion of segments that convey a complete meaning. A segment is considered complete if it expresses a coherent and relevant idea about the feedback. 
    \begin{multline*}
        \text{Completeness} = \\\frac{{Number\ of\ complete\ segments}}{{Total\ number\ of\ segments\ generated}}
    \end{multline*}
    \item \textbf{Correctness/Precision}: This metric measures the proportion of segments that match both the text segment and the image. A segment is considered correct (relevant) if it accurately reflects the information and sentiment from both sources. 
    \begin{multline*}
        \text{Correctness} = \\\frac{{Number\ of\ correct\ segments}}{{Total\ number\ of\ segments\ generated}}
    \end{multline*}
   \item \textbf{Recall}: This metric measures how many relevant (to the image) verbatims are extracted among all the relevant verbatims present in the feedback.
   \begin{multline*} 
   \text{Recall} = \\\frac{{Number\ of\ relevant\ verbatims}}{{Total\ number\ of\ relevant\ verbatims}} 
   \end{multline*}
\end{enumerate}

\onecolumn
\subsection{Training Data}
\label{sec:training_data}
\small
\begin{longtable}[c]{m{3.8cm} m{2.2cm} m{1.9cm} m{3cm} m{1.4cm}}
\hline
\multicolumn{2}{c}{\textbf{Input}} & \multicolumn{3}{c}{\textbf{Output}}                                                            \\
\hline
\endfirsthead
\endhead
%
  \multicolumn{1}{c}{\textbf{Feedback Text}} &
  \multicolumn{1}{c}{\textbf{Feedback Image}} &
  \multicolumn{1}{c}{\textbf{Feedback Image}} &
  \multicolumn{1}{c}{\textbf{Verbatim}} &
  \multicolumn{1}{c}{\textbf{Similarity Score}}\\ 

\hline


\multirow{6}{3.8cm}{\parbox{1\linewidth}{\vspace{1cm} They don't look nice. It looked nice for a brief period of time; then the finish came off, that fadeout the pleated color and became brownish. The one I'd purchased elsewhere before lasted years, looked nice and was retired only because the "stone" was lost.}} &
  \multirow{6}{2.2cm}{\parbox{1\linewidth}{\vspace{2cm} {\includegraphics[width=0.9\linewidth, valign=B]{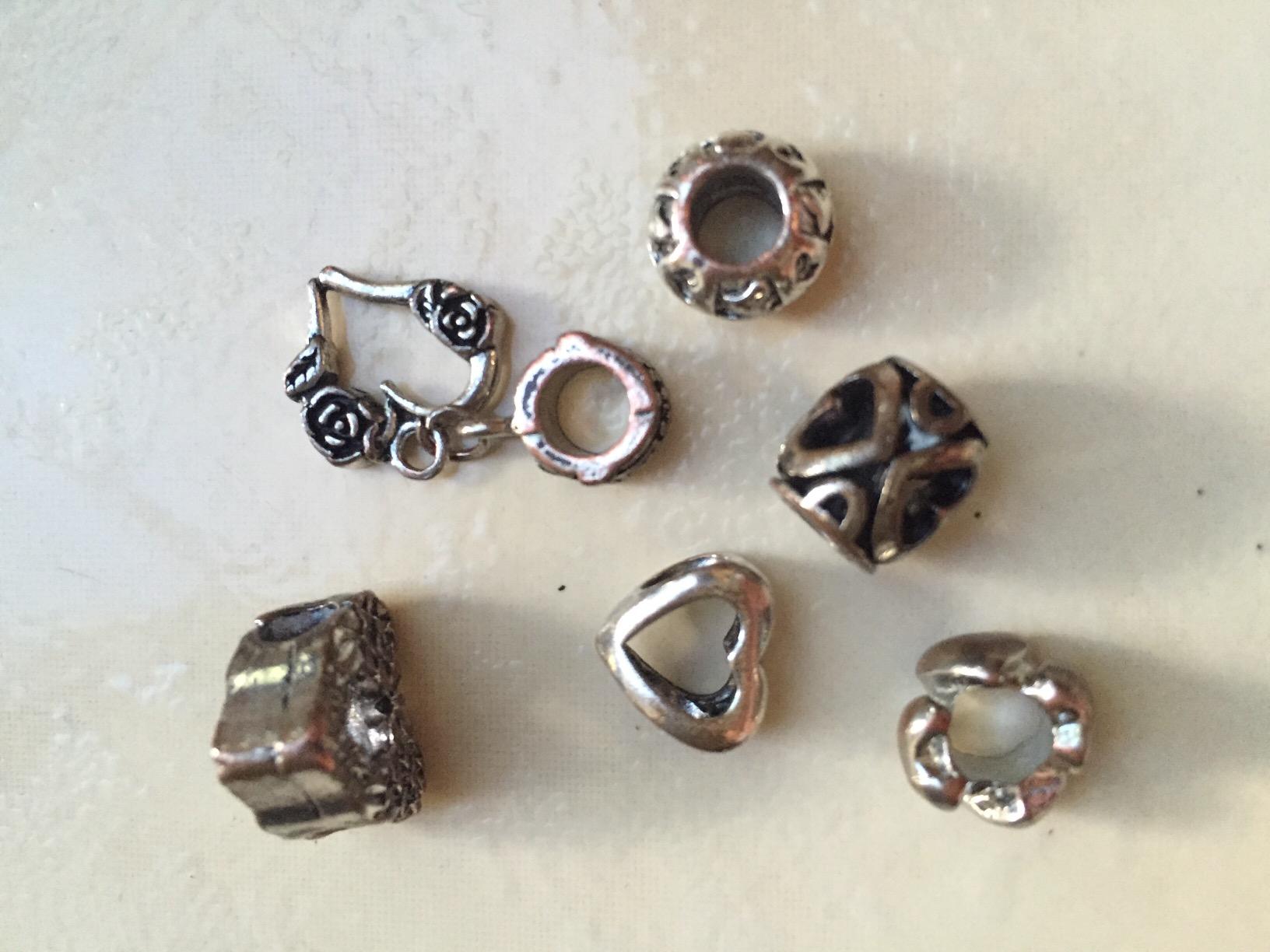}}}}
   &
   {\parbox{1\linewidth}{\vspace{0.2mm} {\includegraphics[width=0.6\linewidth]{image_1}}}}
   &
  They don't look nice &
  0.19 \\
  \cline{3-5}
            &               & {\parbox{1\linewidth}{\vspace{0.2mm} {\includegraphics[width=0.6\linewidth]{image_1}}}}      & It looked nice for a brief period of time                     & 0.12             \\
\cline{3-5}
            &               & {\parbox{1\linewidth}{\vspace{0.2mm} {\includegraphics[width=0.6\linewidth]{image_1}}}}       & then the finish came off                                      & 0.29             \\
\cline{3-5}
            &               & {\parbox{1\linewidth}{\vspace{0.2mm} {\includegraphics[width=0.6\linewidth]{image_1}}}}        & fadeout the pleated color and became brownish                          & 0.31             \\
\cline{3-5}
            &               & {\parbox{1\linewidth}{\vspace{0.2mm} {\includegraphics[width=0.6\linewidth]{image_1}}}}      & The one I'd purchased elsewhere before lasted years           & 0.13             \\
\cline{3-5}
            &               & {\parbox{1\linewidth}{\vspace{0.2mm} {\includegraphics[width=0.6\linewidth]{image_1}}}}        & looked nice and was retired only because the "stone" was lost & 0.21   \\    
\hline
\caption{Training Data}
\label{tab:training_data}\\
\end{longtable}
\normalsize

\subsection{Experimented Prompts}
\label{sec:experimented_prompts}

\begin{longtable}[c]{m{1.2cm}  m{6cm}  m{7.5cm}}
\hline
\multicolumn{1}{m{1.2cm}}{\textbf{Task}} & \multicolumn{1}{c}{\textbf{Input Prompt}} & \multicolumn{1}{c}{\textbf{Target}} \\
\endfirsthead
\endhead
\hline
CSECS &
  Extract all the verbatim and confidence score of each matching with image? Feedback: <feedback text> &

\emph{<verbatim 1>; <confidence 1> | <verbatim 2>; <confidence 2> }
 \\
\cline{1-3} 
MSECS &
  Extract all the verbatim and confidence score of each matching with image?  Feedback: <feedback text> &

\emph{<matching verbatim 1>; <confidence 1> | \, \, \, \, <matching verbatim 2>; <confidence 2>} 
\\
  \cline{1-3}
MSE &
  What is the verbatim matching with the image? Feedback: <feedback text> &
\emph{<matching verbatim 1> | <matching verbatim 2> }
\\
  \hline
\caption{Experimented Prompts}
\label{tab:experimented_prompts}\\
\end{longtable}

\clearpage
\subsection{MINE: Sample Predictions}
\label{sec:sample_predictions}

\begin{longtable}[c]{m{7.4cm}  m{2.3cm} | m{2.3cm}  m{2.5cm}}
\hline
\multicolumn{2}{c|}{\textbf{Model Input (Multi-modal Feedback)}} & \multicolumn{2}{c}{\textbf{Model Output}} \\ \hline
\endfirsthead
\multicolumn{4}{c}%
{{\bfseries Table \thetable\ continued from previous page}} \\
\hline
\multicolumn{2}{c}{\textbf{Model Input (Multi-modal Feedback)}} & \multicolumn{2}{c}{\textbf{Model Output}} \\ \hline
\multicolumn{1}{c}{\textbf{Text}} & \textbf{Image} & \textbf{Image} & \textbf{Verbatim} \\ \hline
\endhead
\multicolumn{1}{c}{\textbf{Text}} & \textbf{Image} & \textbf{Image} & \textbf{Verbatim} \\ \hline
\multirow{2}{7.4cm}{Paint is outside the lines and it really looks sloppy. So, i really dont like it. It looks so bad. I would not order this it is product. And it was rolled up like a newspaper. Complete and total waste. The paint is pasted on the nose as well. } & {\includegraphics[width=0.65\linewidth]{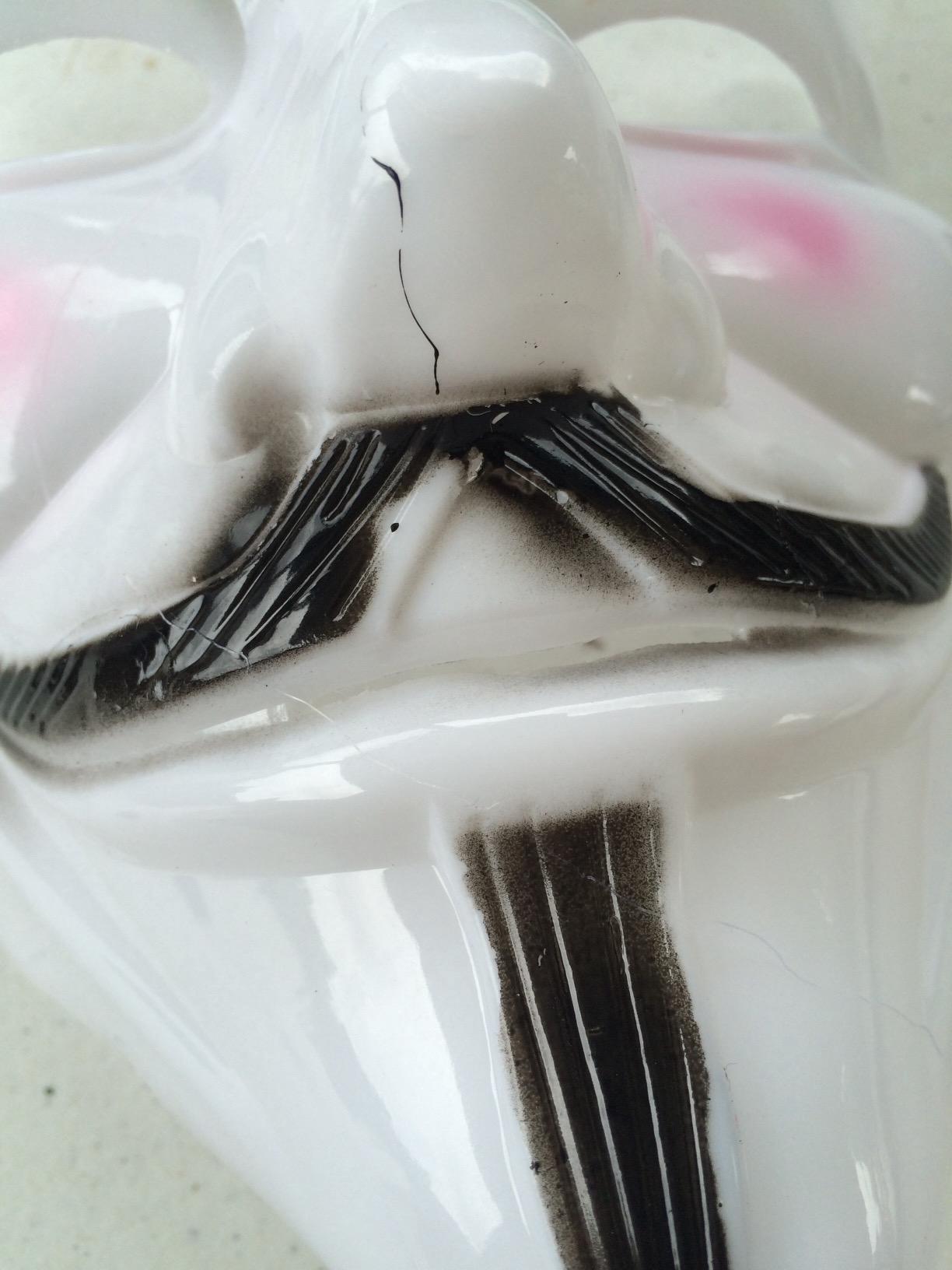}} & {\includegraphics[width=0.5\linewidth]{image_2}} & The paint is pasted on the nose as well \\ \cline{3-4} 
 & {\includegraphics[width=0.65\linewidth]{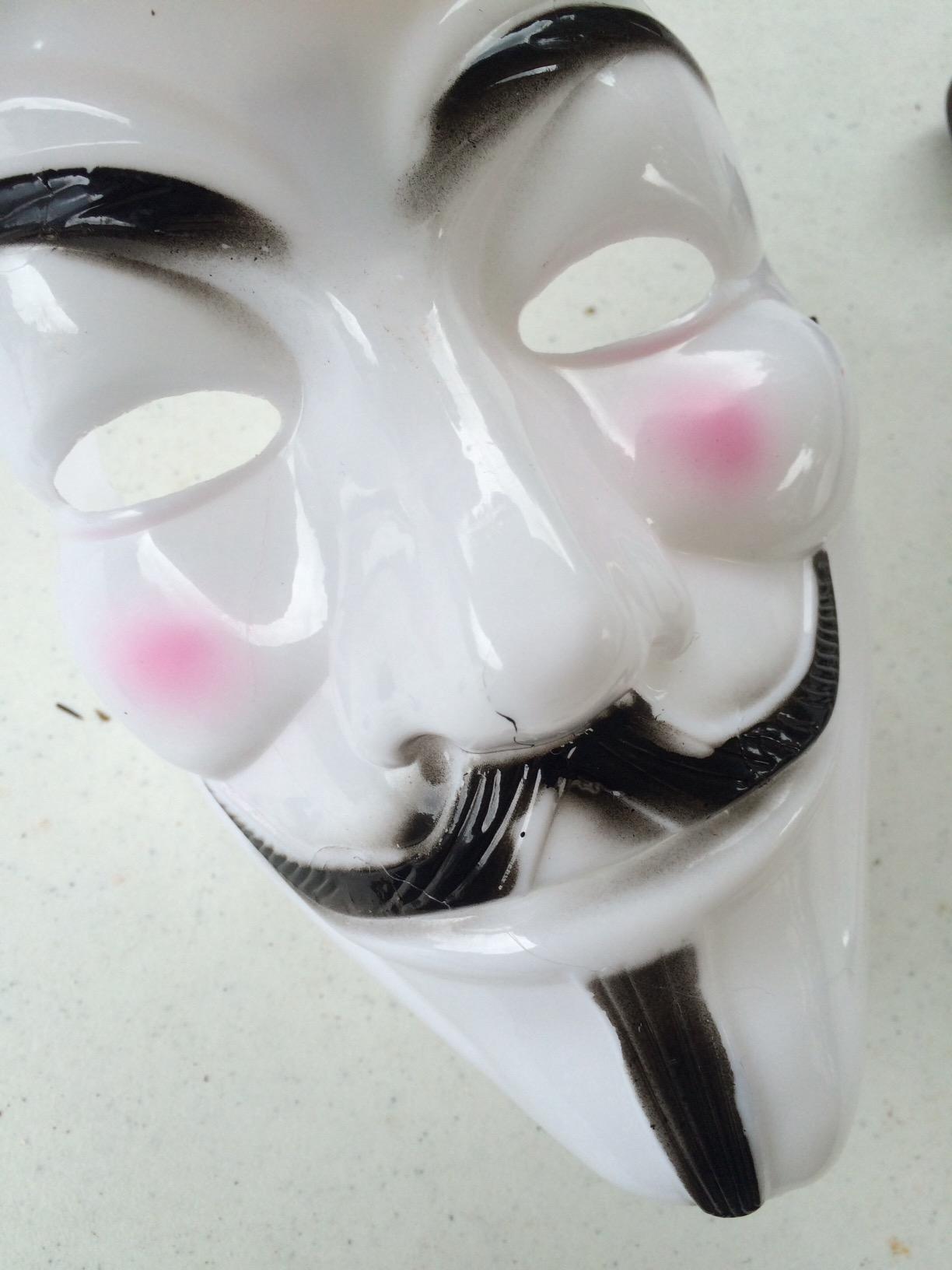}} & {\includegraphics[width=0.5\linewidth]{image_3}} & Paint is outside the lines and it really looks sloppy \\ \hline

\multirow{3}{7.4cm}{\parbox[][20cm][s]{1\linewidth}{\vspace{-2mm}I was disappointed. The shoulder straps did not have anything to secure the loose strap so you essentially have a loose strap getting in the way. This makes the purse look really cheap. Inside, the zipper is misaligned so when it is zipped, you see a gap. When I try to rezip, it doesn't correct itself. Lastly, the inside pocket must have been oversewn, because when you open the pocket, you can see the bright green threading. Lastly, the gold-plated clasps that connect the purse and shoulder straps together were also not aligned.}} & {\includegraphics[width=0.65\linewidth]{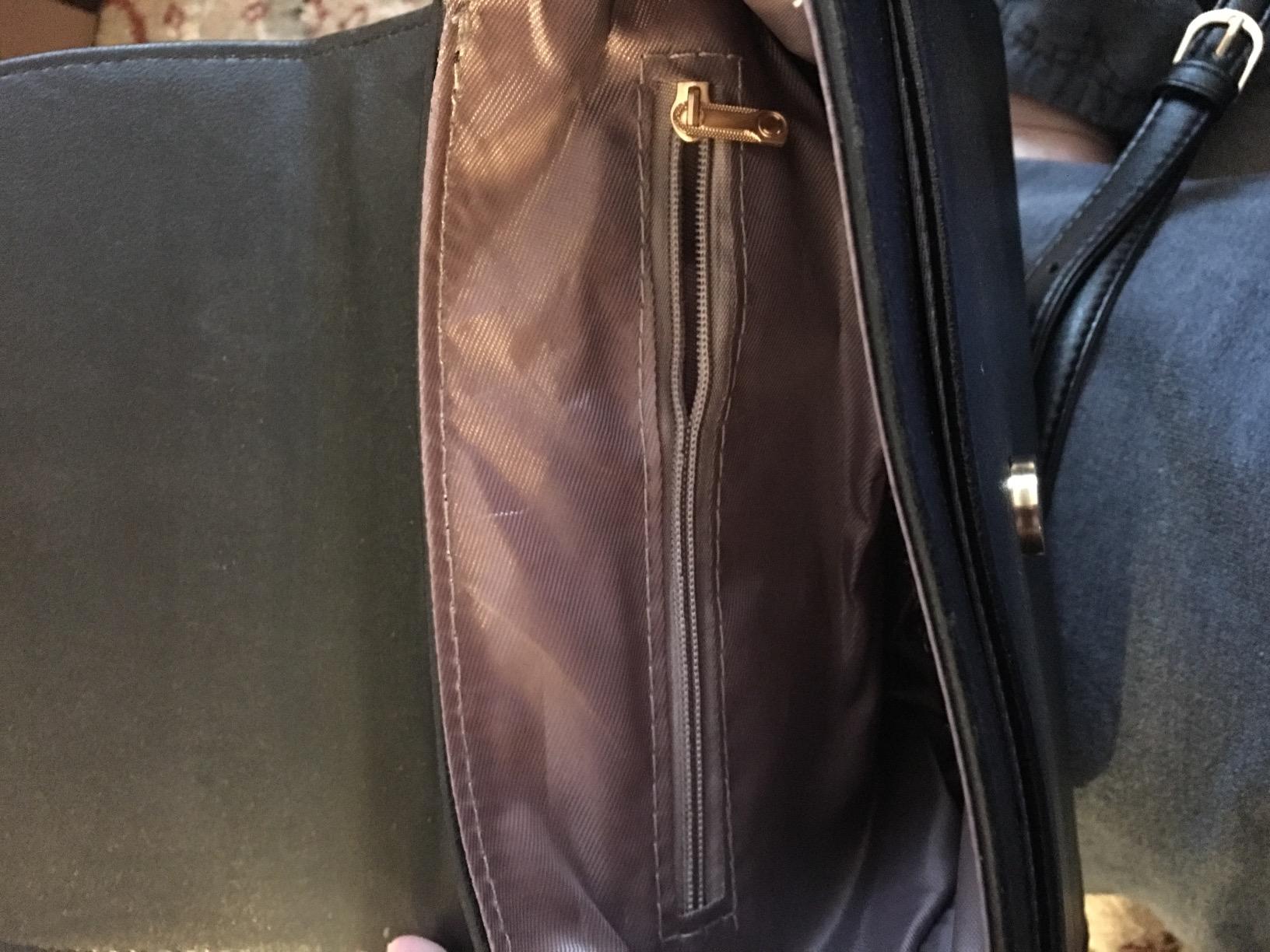}} & {\includegraphics[width=0.5\linewidth]{image_4}} & the zipper is misaligned \\ \cline{3-4} 
 & {\includegraphics[width=0.65\linewidth]{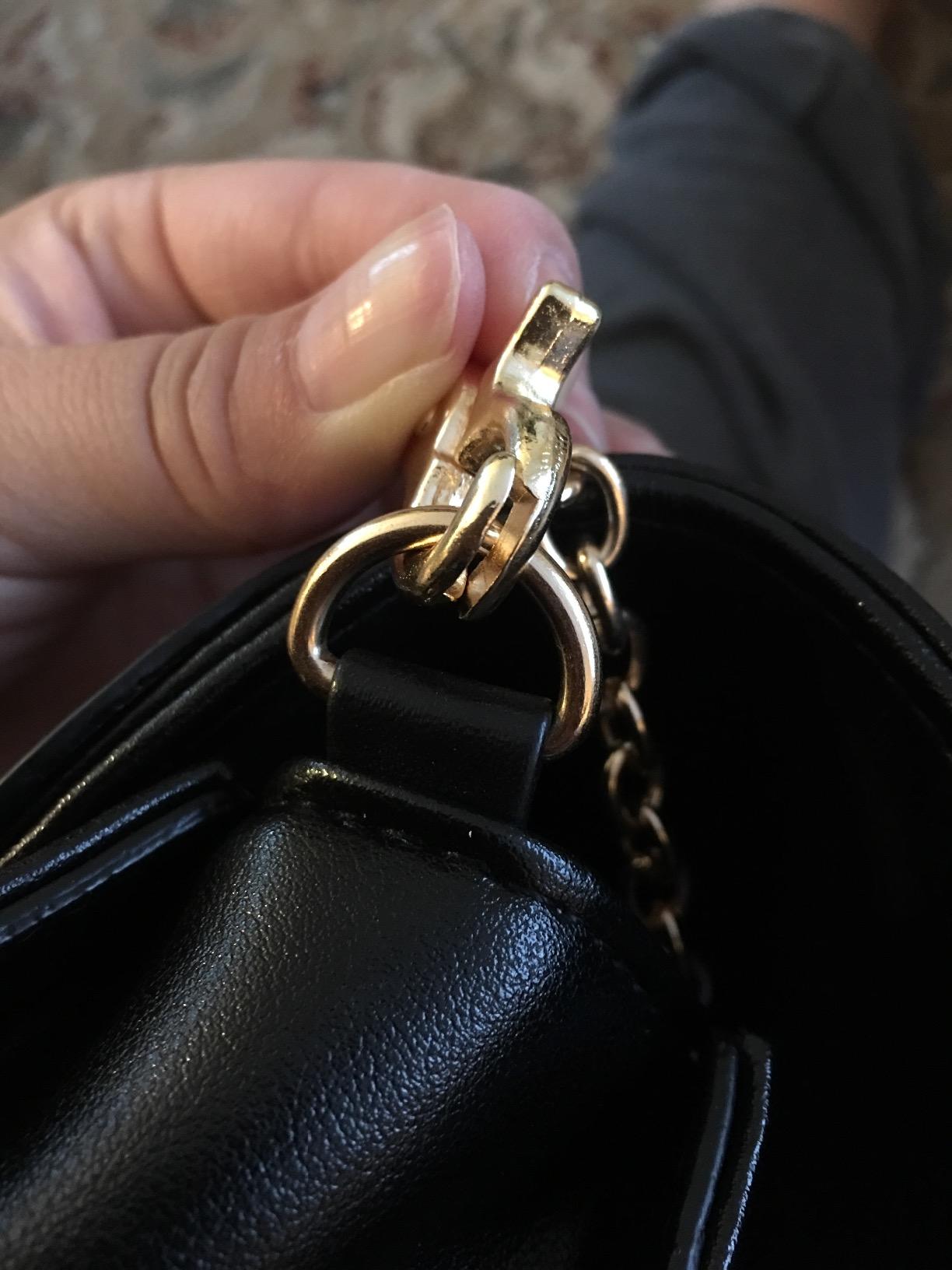}} & {\includegraphics[width=0.5\linewidth]{image_5}} & shoulder straps together were also not aligned \\ \cline{3-4} 
 & {\includegraphics[width=0.65\linewidth]{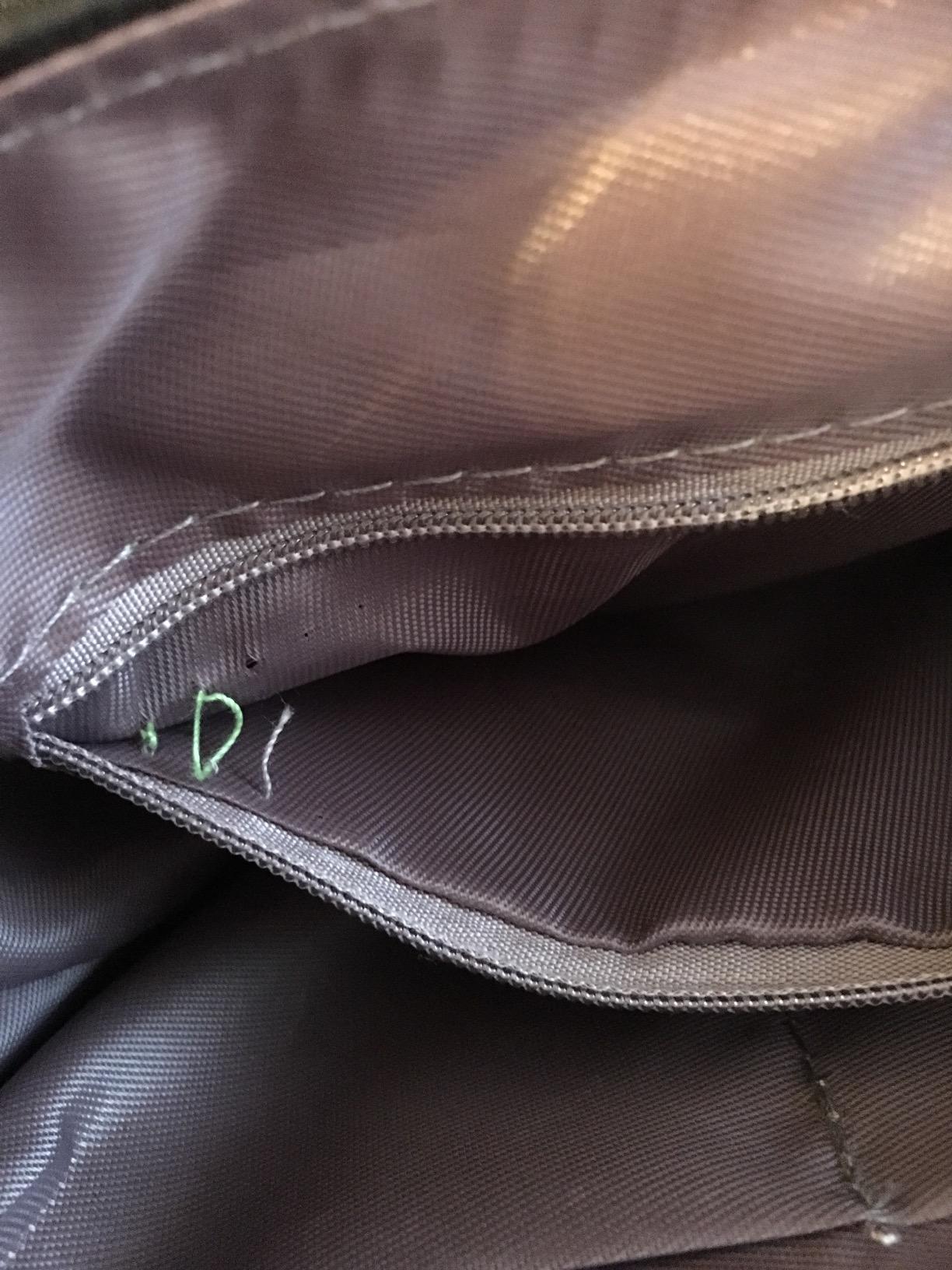}} & {\includegraphics[width=0.5\linewidth]{image_6}} & inside pocket must have been oversewn \\
 \hline
\multirow{3}{7.4cm}{\parbox{1\linewidth}{\vspace{1mm}{I wish the product to be good and expecting for it. Got the product just by today morning. It never met my expectations and I am totally disappointed!. This set is really small. The shakers don't fit in the holder and the holder is pathetic! Not made well at all!!!}}} & {\includegraphics[width=0.65\linewidth]{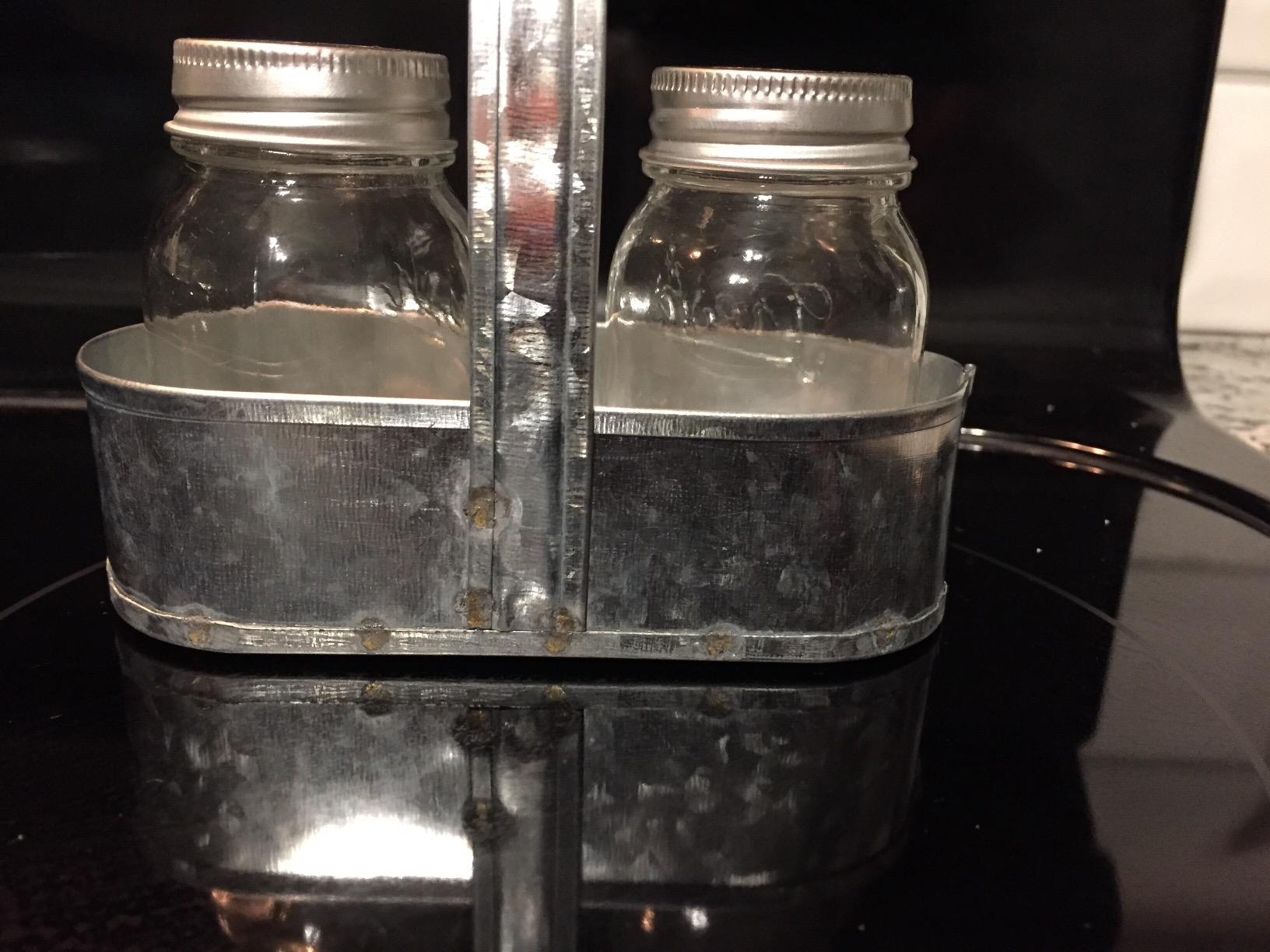}} & \multirow{3}{2.3cm}{\parbox{1\linewidth}{\vspace{1.5cm}{\includegraphics[width=0.5\linewidth]{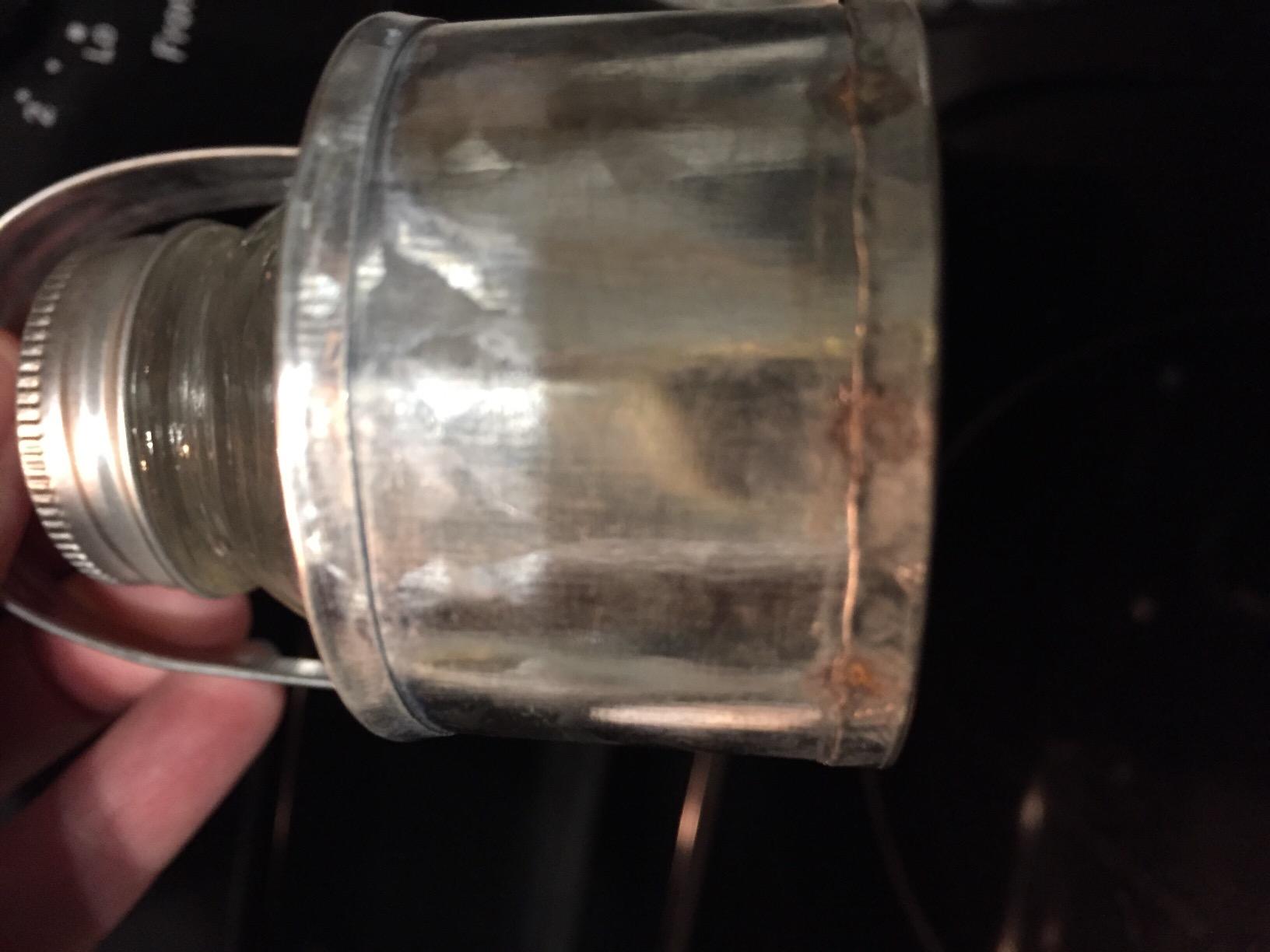}}}} & \multirow{3}{2.5cm}{\parbox{1\linewidth}{\vspace{1.6cm}{Not made well at all}}} \\ 
 & {\includegraphics[width=0.65\linewidth]{image_8}} &  &  \\ 
 & {\includegraphics[width=0.65\linewidth]{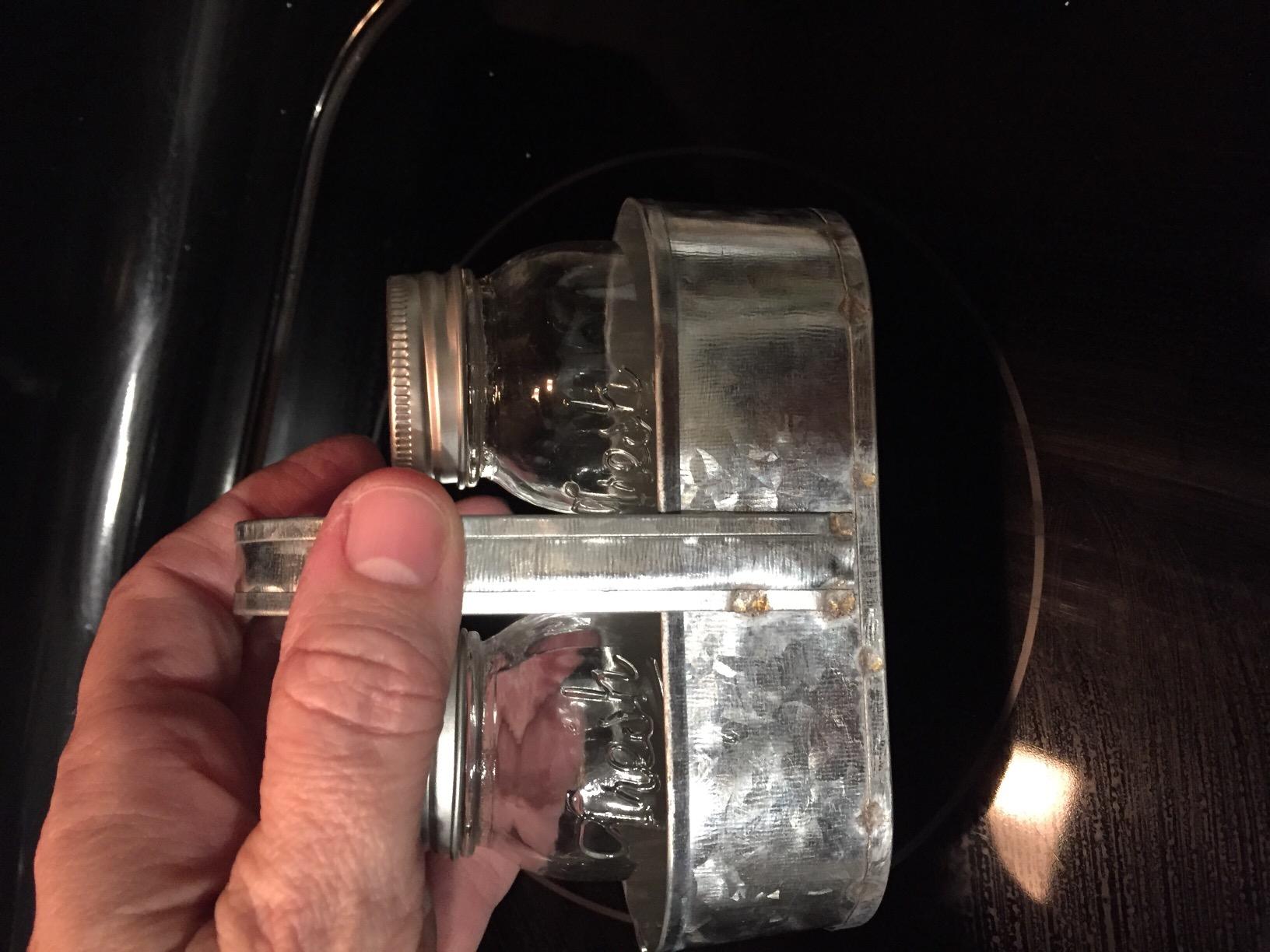}} &  &  \\ \hline
 \multirow{2}{7.4cm}{\parbox{1\linewidth}{\vspace{-5mm}{I've had these boots for a few months and only wear them when I ride (less than twice a week). For no reason at all the heel on the right boot fell off (in half). Looks like the heel is made of particle cardboard. I would have expected that boots in this price range would not fall apart under light use in just a few months! What a rip off!}}} & {\includegraphics[width=0.65\linewidth, height=16mm]{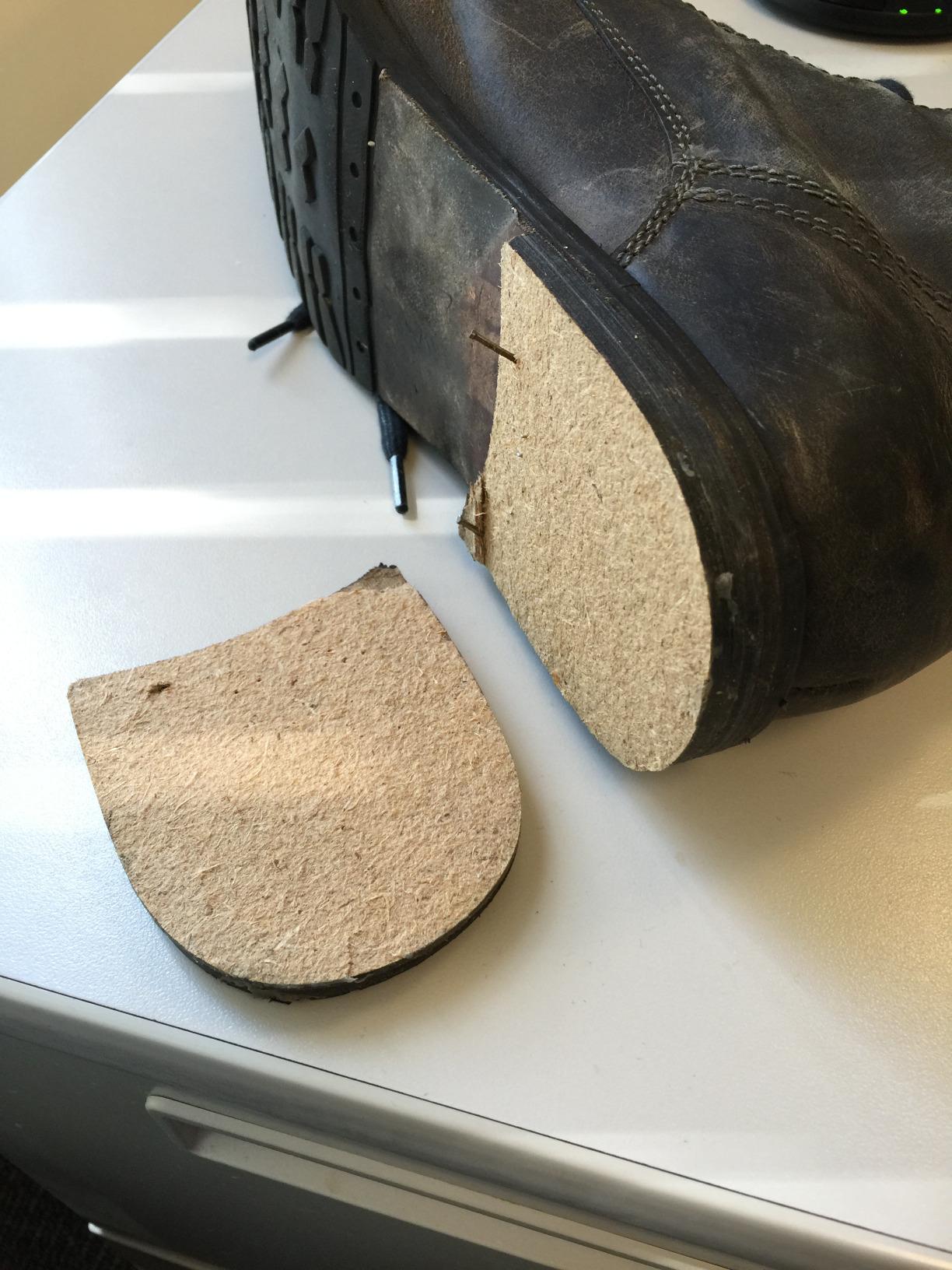}} & {\includegraphics[width=0.5\linewidth, height=12mm]{image_10}} & heel on the right boot fell off, \\ \cline{3-4}
 & {\includegraphics[width=0.65\linewidth]{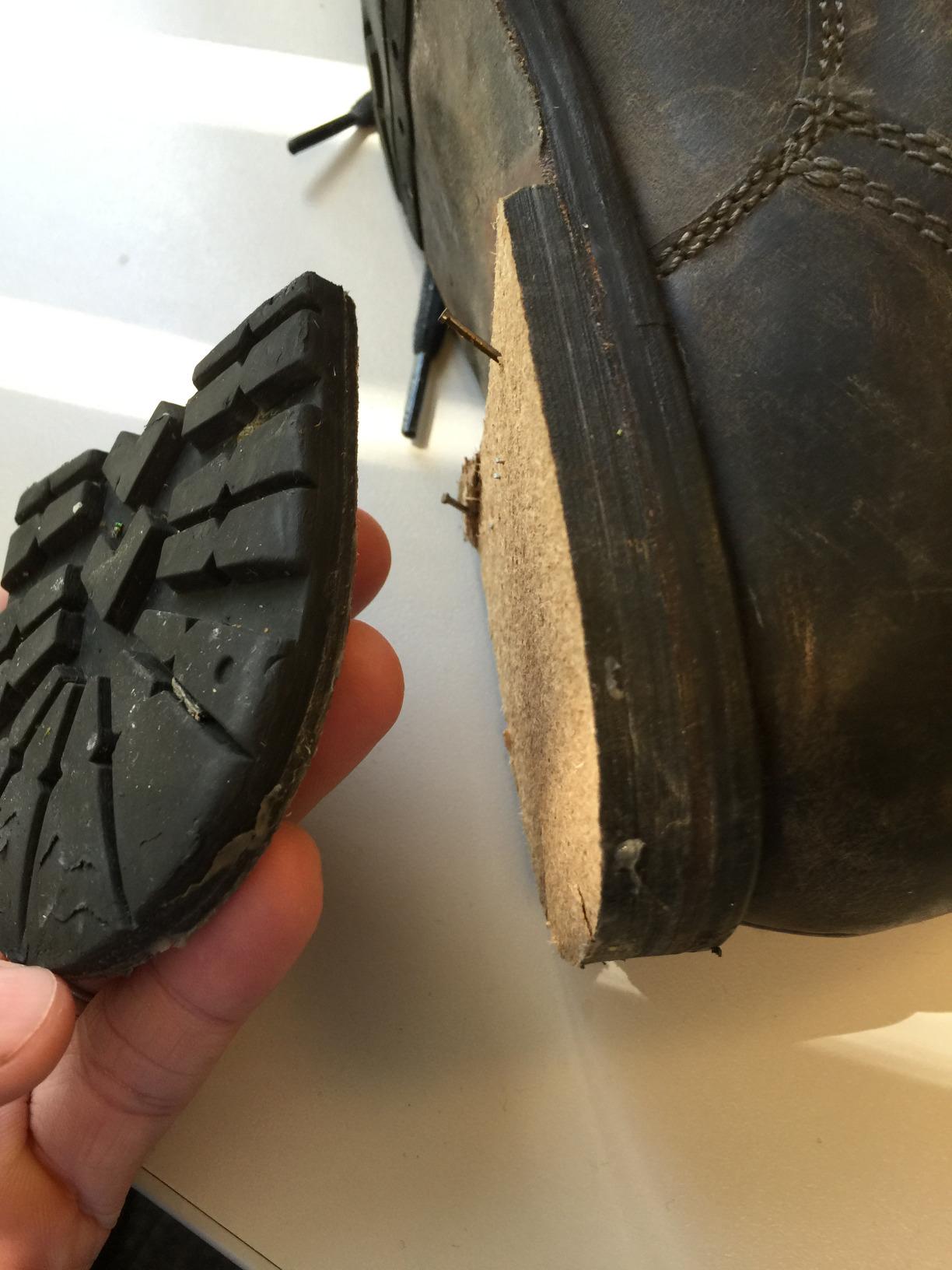}} & {\includegraphics[width=0.5\linewidth]{image_11}} & fall apart under light use. 
 \\ \hline
 \multirow{3}{7.4cm}{\parbox{1\linewidth}{\vspace{-4mm}{I was very pleased with the print and very excited to hang these up. We had a room that was lonely and in need of some fun print, these looked like the perfect fit. I was very disappointed once they were hung. The packaging stated 95" length; what I received was one of 88" and the other of 92". Although the print was nice, what can one do with them? very disappointed in this purchase.
}}} & {\includegraphics[width=0.65\linewidth]{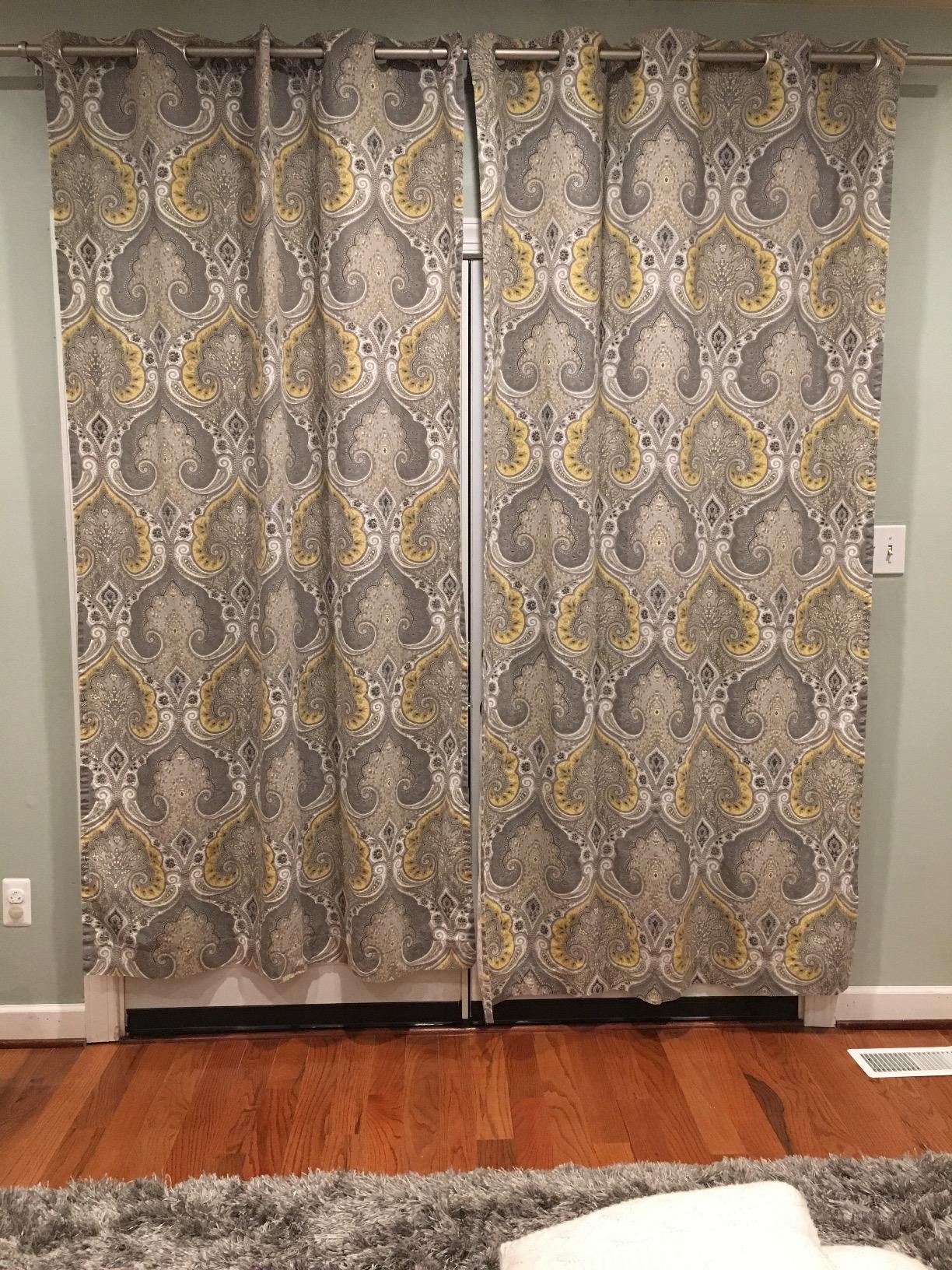}} & {\includegraphics[width=0.5\linewidth]{image_12}} & received was one of 88" and the other of 92" \\ \cline{3-4}
 & {\includegraphics[width=0.65\linewidth]{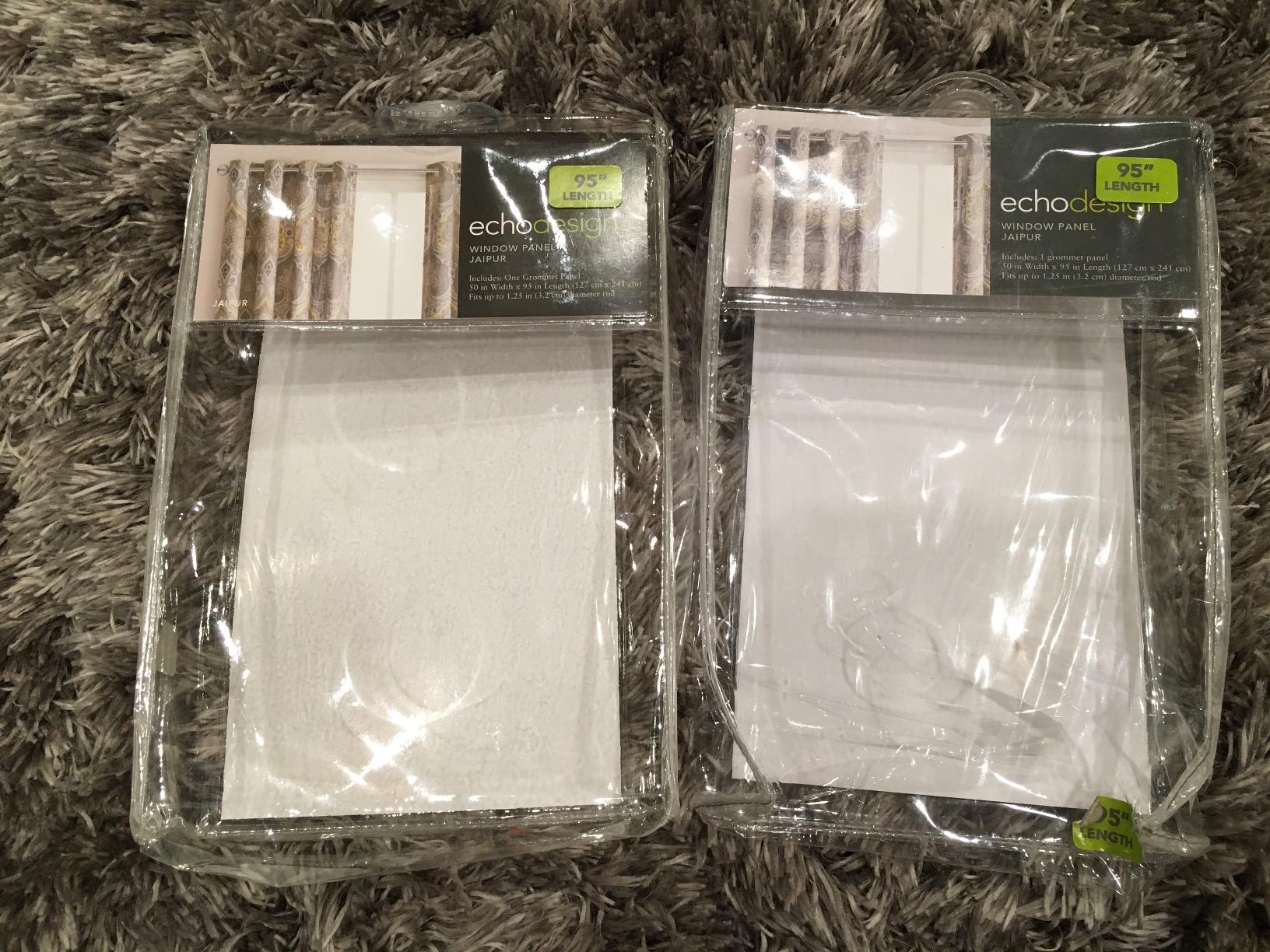}} & \multirow{2}{2.3cm}{\parbox{1\linewidth}{\vspace{5mm}{\includegraphics[width=0.5\linewidth]{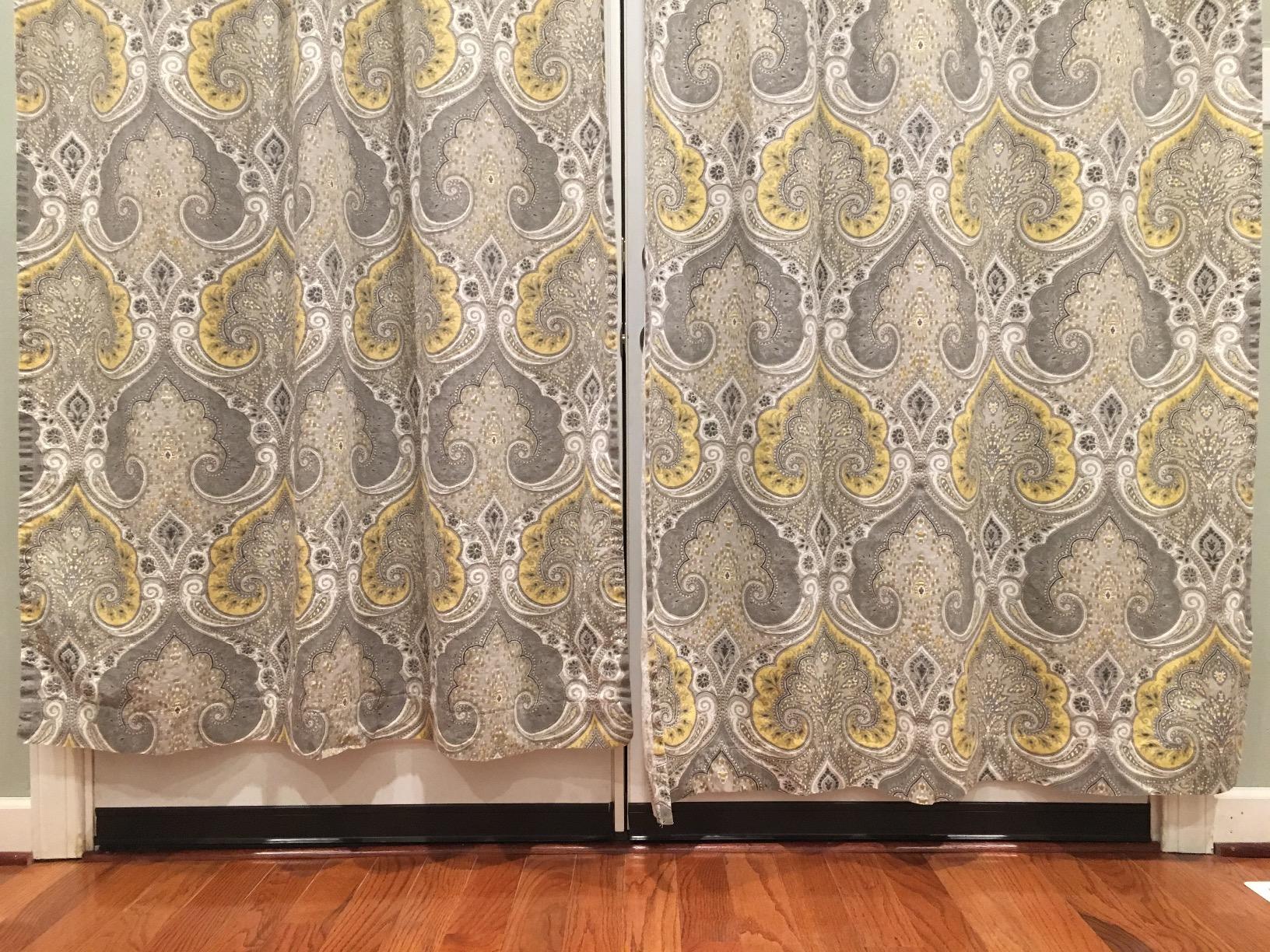}}}} & \multicolumn{1}{c}{\multirow{2}{2.5cm}{\parbox{1\linewidth}{\vspace{1mm}{received was one of 88" and the other of 92"}}}} \\ 
 & {\includegraphics[width=0.65\linewidth]{image_14}} &  &  \multicolumn{1}{c}{} \\ \hline
 \multirow{3}{7.4cm}{\parbox{1\linewidth}{\vspace{-3mm}{The first one I ordered had a big chip on one of the small pieces and I found the loose chip in the package too. They sent a replacement pretty quick, but the second one was even worse - multiple cracks, uneven finish and one big problem - one of the main middle pieces did not fit the connecting piece.  The hole was not big enough.  This was a gift, and I felt so embarrassed.}}} & {\includegraphics[width=0.7\linewidth]{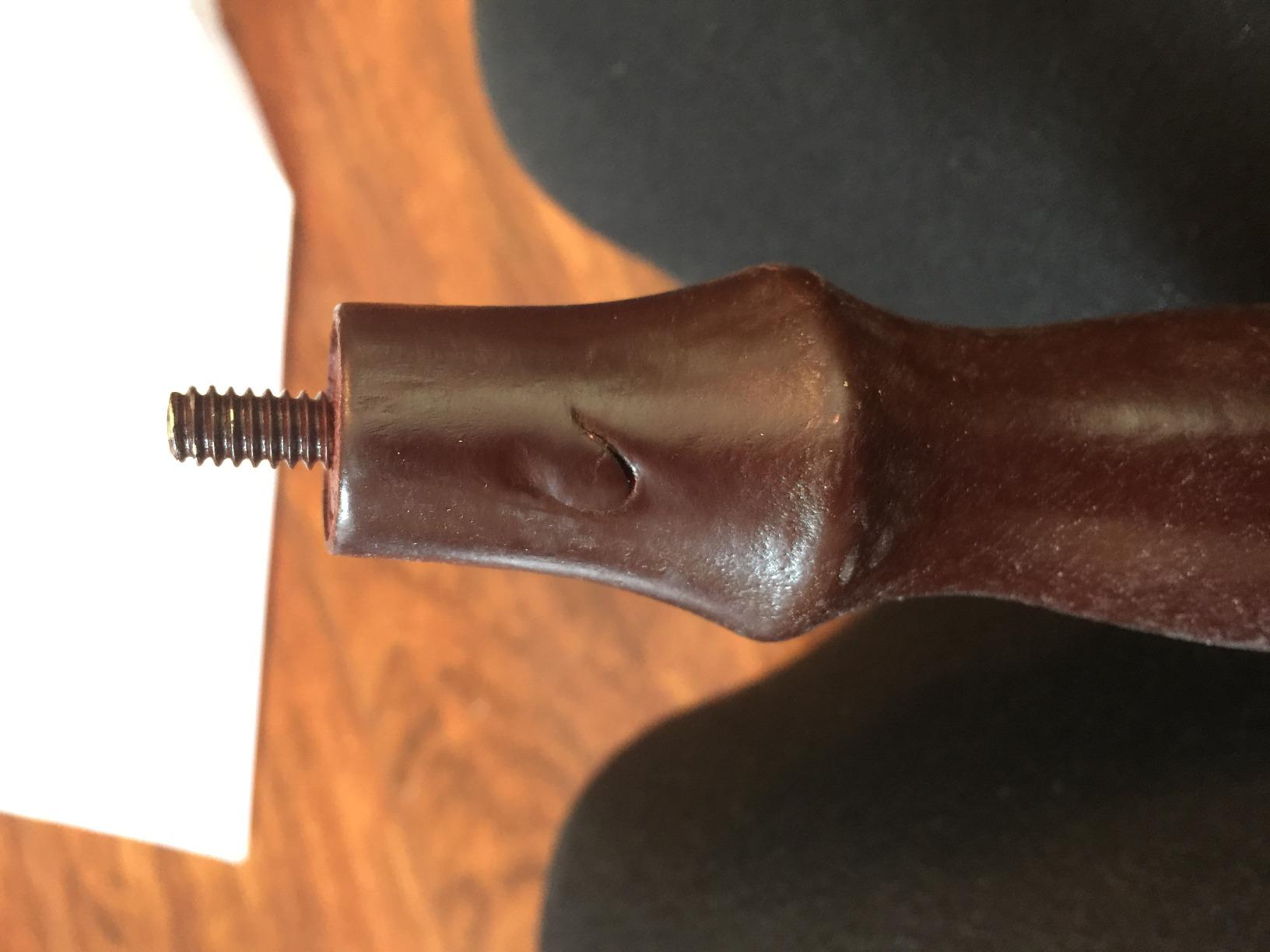}} &  {\includegraphics[width=0.5\linewidth]{image_15}} &  even worse - multiple cracks \\ \cline{3-4}
 & {\includegraphics[width=0.7\linewidth]{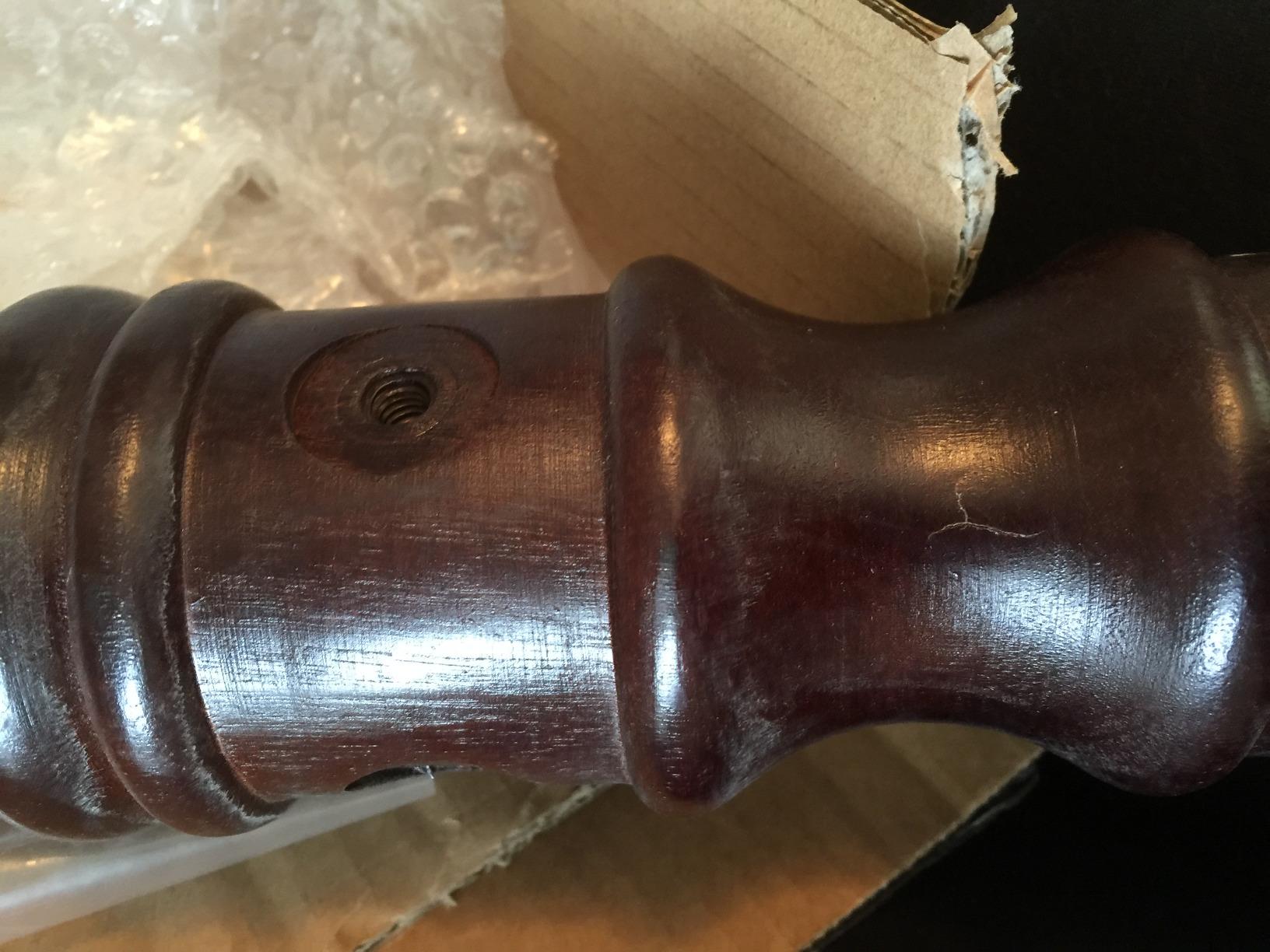}} & \multirow{2}{2.3cm}{\parbox{1\linewidth}{\vspace{6mm}{\includegraphics[width=0.5\linewidth]{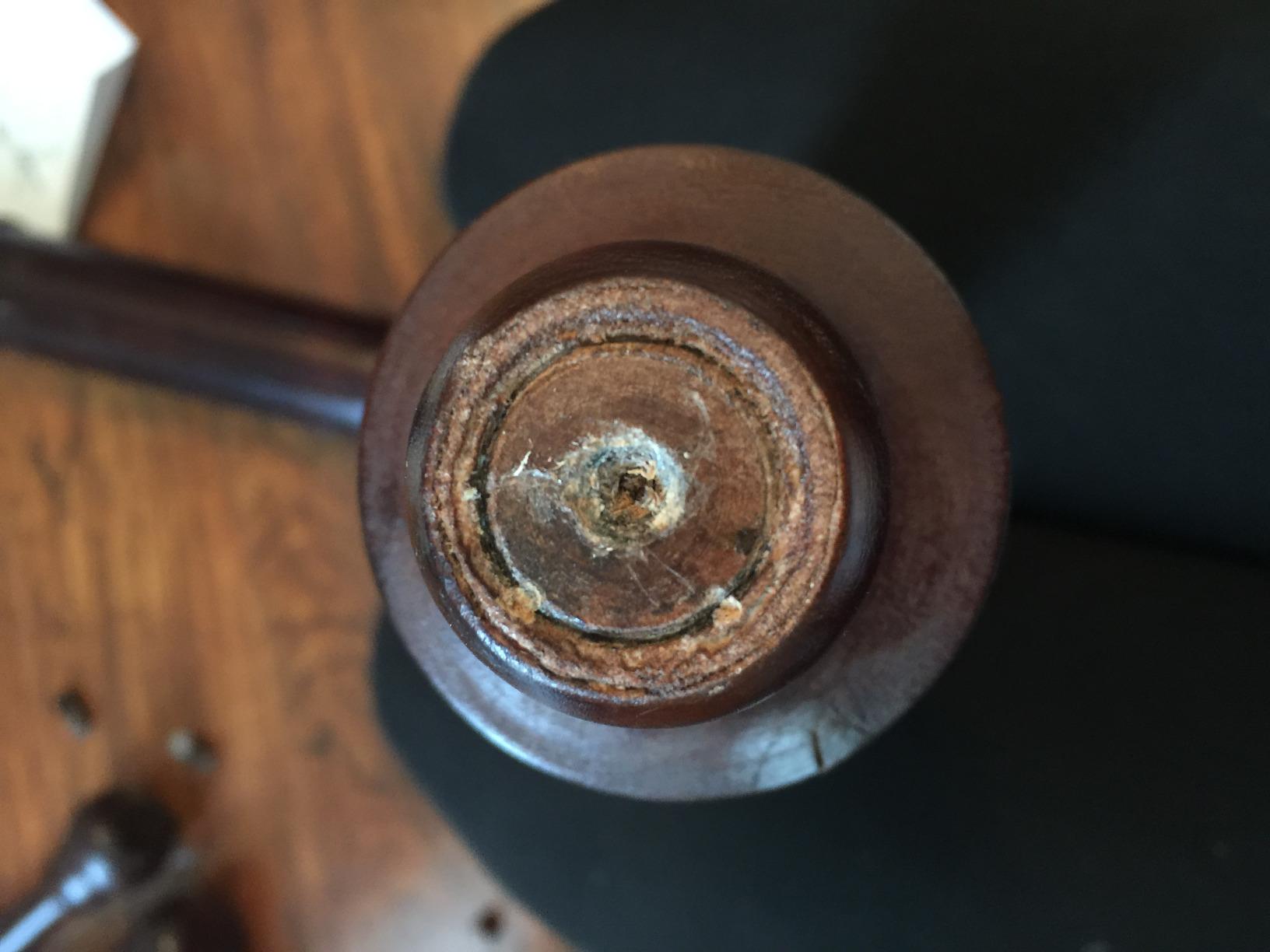}}}} & \multicolumn{1}{c}{\multirow{2}{2.5cm}{\parbox{1\linewidth}{\vspace{4mm}{ hole was not big enough}}}} \\ 
 & {\includegraphics[width=0.7\linewidth]{image_17}} &  & \multicolumn{1}{c}{} \\
 \hline
\multirow{4}{7.4cm}{\parbox{1\linewidth}{\vspace{-1mm}{Very disappointed in this ring. I have worn it most every day, taking it off for showers, for the past couple of months and it's already tarnishing, and it looks like the silver is coming off of the band. Looks more like it's silver plated instead of actual sterling silver. I have owned SS and Black Hills Gold Jewelry for years and I am careful not to expose my jewelry to harsh chemicals, etc.}}} & {\includegraphics[width=0.7\linewidth]{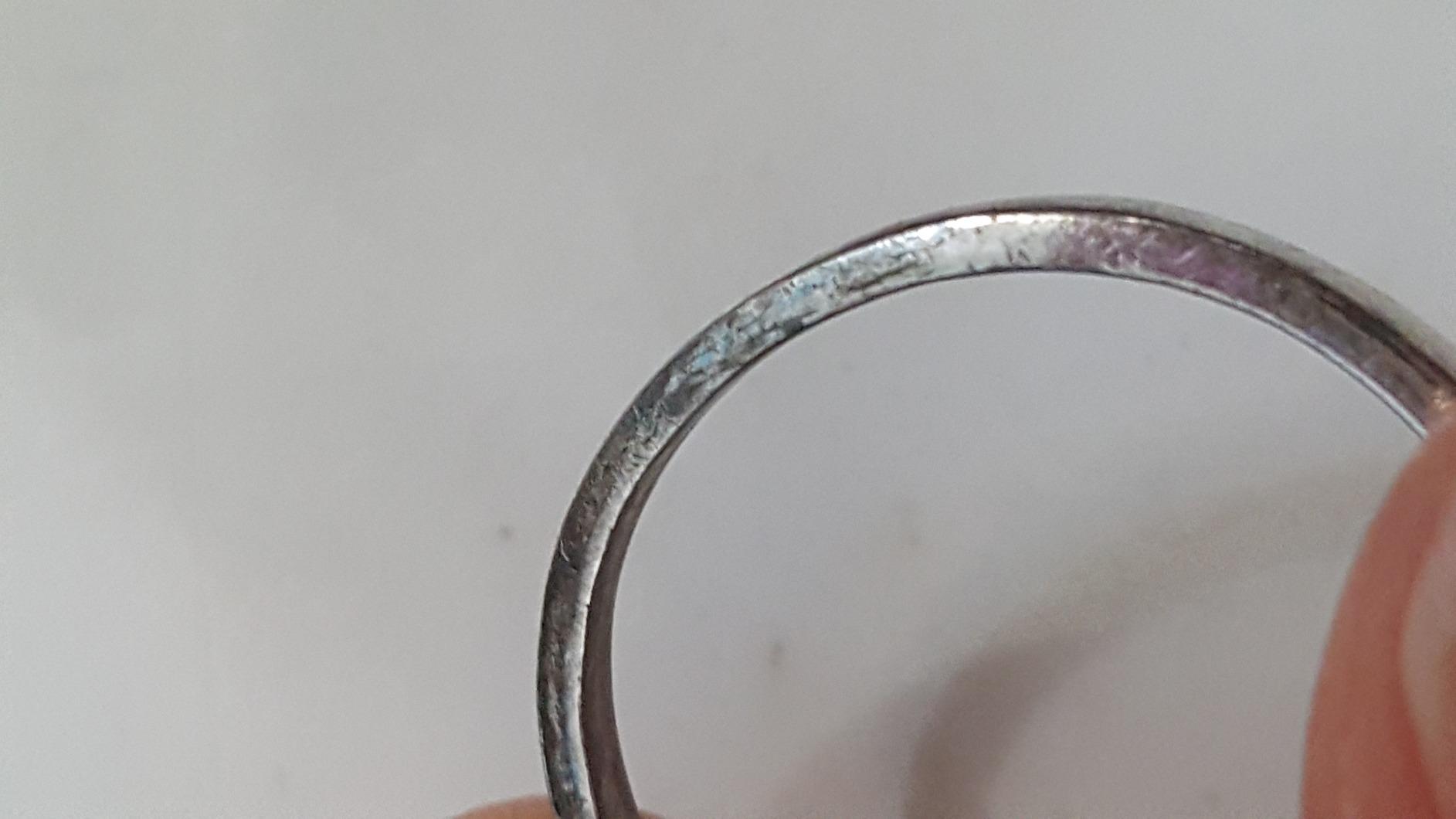}} & \multirow{2}{2.3cm}{\parbox{1\linewidth}{\vspace{7mm}{\includegraphics[width=0.5\linewidth]{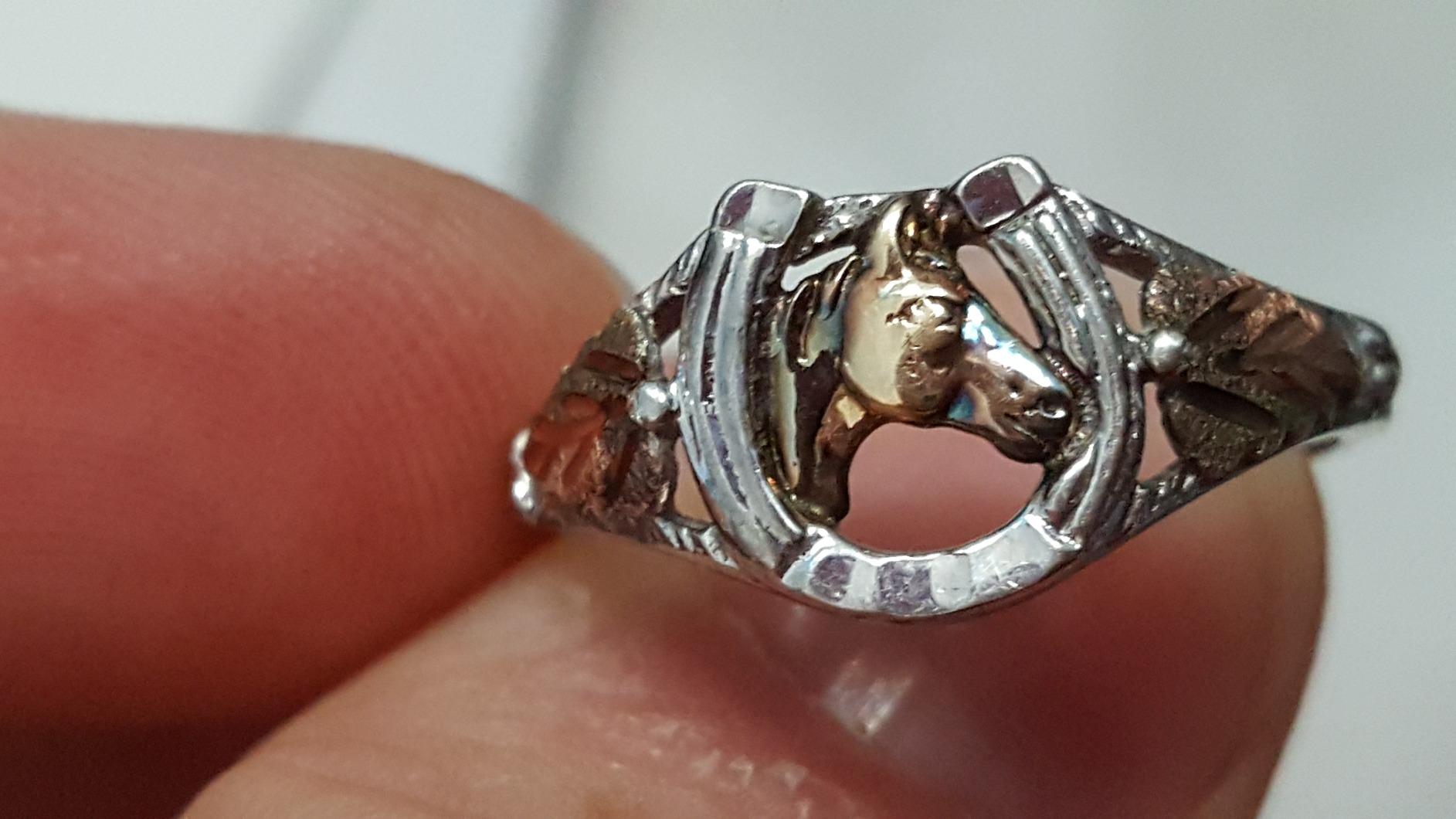}}}} & \multirow{2}{2.5cm}{\parbox{1\linewidth}{\vspace{6mm}{it's already tarnishing}}} \\ 
 & {\includegraphics[width=0.7\linewidth]{image_19}} &  &  \\ \cline{3-4} 
 & {\includegraphics[width=0.7\linewidth]{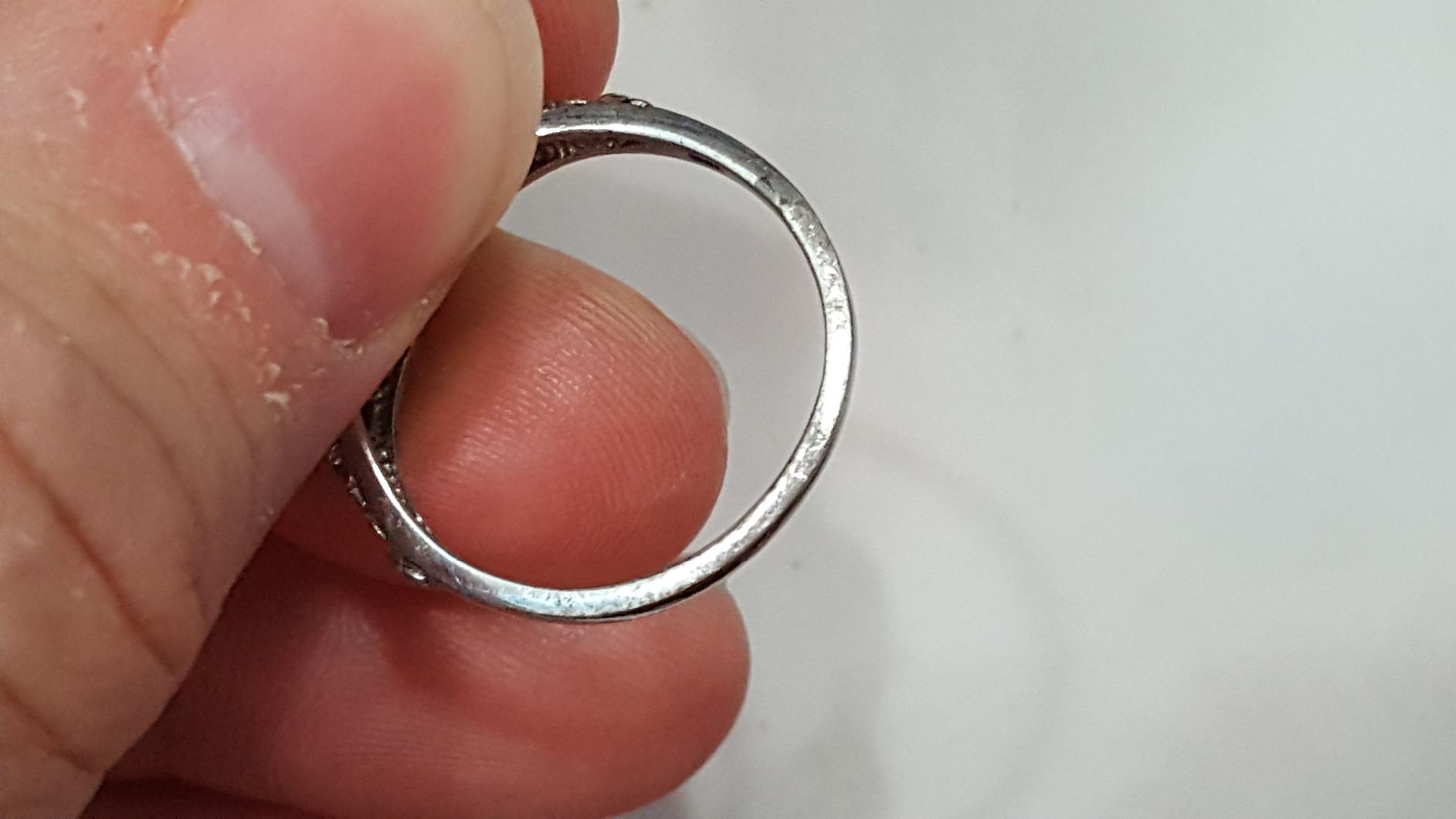}} & \multirow{2}{2.3cm}{\parbox{1\linewidth}{\vspace{7mm}{\includegraphics[width=0.5\linewidth]{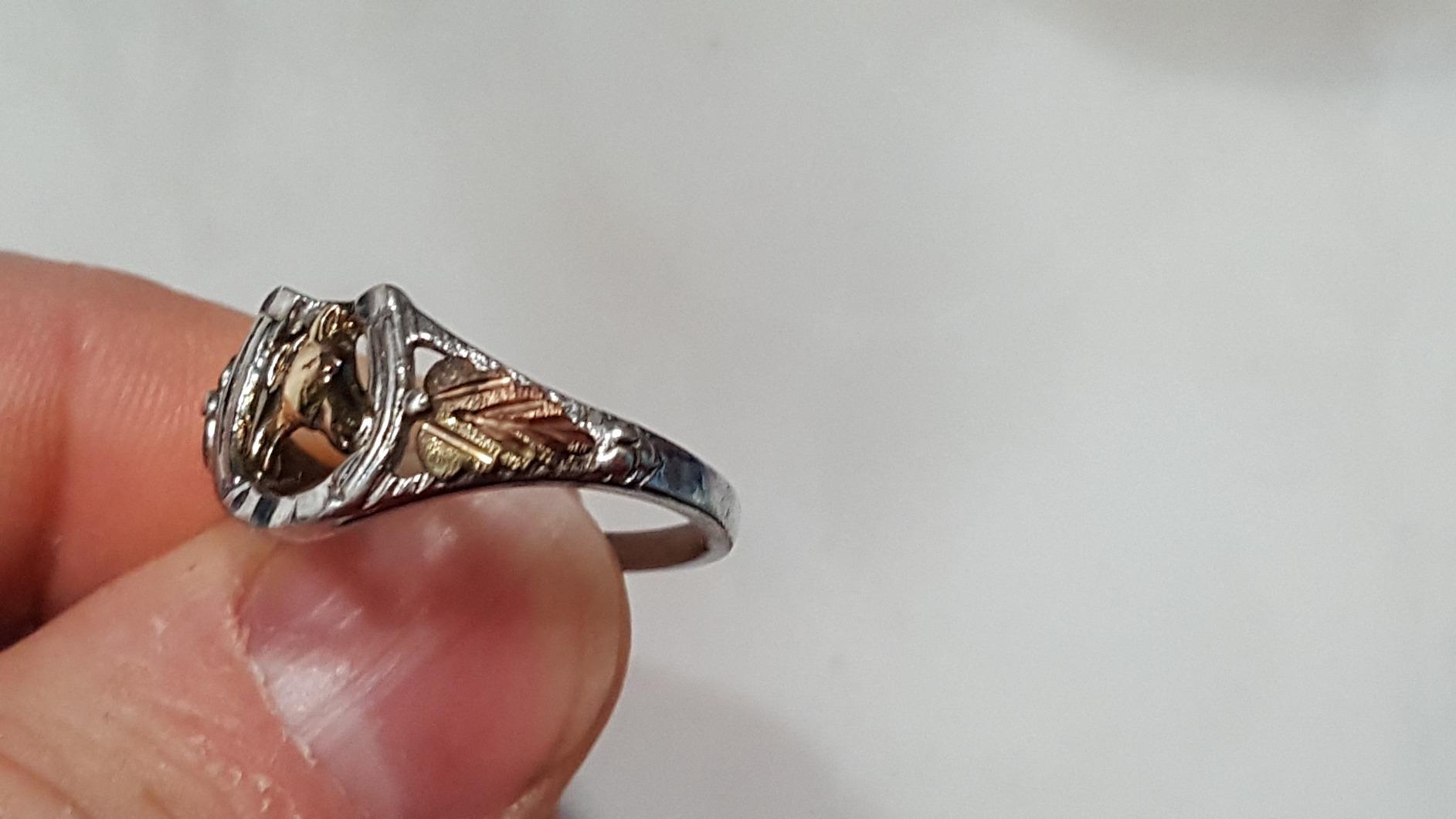}}}} & \multirow{2}{2.5cm}{\parbox{1\linewidth}{\vspace{6mm}{it's already tarnishing}}} \\ 
 & {\includegraphics[width=0.7\linewidth]{image_21}} &  &  \\ \hline
\caption{Sample Predictions}
\label{tab:sample_prediction}\\
\end{longtable}

\twocolumn 

\end{document}